\documentclass[a4paper]{article}

\usepackage[margin=1in]{geometry} 
\usepackage{amsmath}
\usepackage{amsthm}
\usepackage{amssymb}
\usepackage{amsfonts} 
\usepackage[utf8]{inputenc} 
\usepackage[T1]{fontenc}    
\usepackage{hyperref}       
\usepackage{url}            
\usepackage{booktabs}       
\usepackage{microtype}      
\usepackage{lipsum}
\usepackage{xfrac}
\usepackage{fancyhdr}       
\usepackage{graphicx}       
\graphicspath{{media_nc/}}     
\usepackage{amsopn,cancel,xcolor}
\usepackage{multirow, booktabs}
\usepackage{subcaption}
\usepackage{placeins}
\usepackage[capitalise,noabbrev]{cleveref}
\usepackage{multirow}
\usepackage{diagbox}

\usepackage[toc,page,titletoc]{appendix}
\Crefname{supp}{Supplement}{Supplements}

\hypersetup{
	unicode,
	pdfauthor={Author One, Author Two, Author Three},
	pdftitle={A simple article template},
	pdfsubject={A simple article template},
	pdfkeywords={article, template, simple},
	pdfproducer={LaTeX},
	pdfcreator={pdflatex}
}

\usepackage[sort&compress,numbers,square]{natbib}
\graphicspath{{media_nc/}}

\title{Real-time Inference and Extrapolation via a Diffusion-inspired Temporal Transformer Operator (DiTTO)}

\author{Oded Ovadia$^1$\thanks{equal contributor.} 
        \and Vivek Oommen$^2$\footnotemark[1]  \and Adar Kahana$^3$ \and Ahmad Peyvan$^3$
         \and Eli Turkel$^1$ \and George Em Karniadakis $^{3,2}$\thanks{Corresponding author.}}

\date{
	$^1$Department of Applied Mathematics, Tel Aviv University, Tel Aviv 69978, Israel \\ \texttt{odedovadia@mail.tau.ac.il, eliturkel@gmail.com}\\%
    $^2$School of Engineering, Brown University, Providence, RI 02912, USA \\ \texttt{\{vivek\_oommen,  george\_karniadakis\}@brown.edu} \\
    $^3$Division of Applied Mathematics, Brown University, Providence, RI 02912, USA \\ 
    \texttt{\{adar\_kahana, ahmad\_peyvan\}@brown.edu}
    \\[2ex]%
}

\begin{document}
\maketitle

\begin{abstract}
    Extrapolation remains a grand challenge in deep neural networks across all application domains. We propose an operator learning method to solve time-dependent partial differential equations (PDEs) continuously and with extrapolation in time without any temporal discretization. The proposed method, named Diffusion-inspired Temporal Transformer Operator (DiTTO), is inspired by latent diffusion models and their conditioning mechanism, which we use to incorporate the temporal evolution of the PDE, in combination with elements from the transformer architecture to improve its capabilities. Upon training, DiTTO can make inferences in real-time. We demonstrate its extrapolation capability on a climate problem by estimating the temperature around the globe for several years, and also in modeling hypersonic flows around a double-cone. We propose different training strategies involving temporal-bundling and sub-sampling and demonstrate performance improvements for several benchmarks, performing extrapolation for long time intervals as well as zero-shot super-resolution in time.
 \href{https://www.youtube.com/playlist?list=PLOH3RQrsYp-6zmthkCiTaUogq2DiyYQnA}{Link to demo}.

\end{abstract}


The field of scientific machine learning (SciML) has been growing rapidly in recent years, and many successful methods for modeling scientific problems using machine learning (ML) methods have been proposed \cite{RAISSI2019686,lu2021learning,li2020fourier,long2018pde,xu2019dl}. 
Many tools designed for standard ML and data science problems can also perform well on SciML tasks. With their unique properties, recent state-of-the-art methods can achieve higher accuracy with fewer data samples, fewer computational resources, more generalization and robustness for a wide range of problems across different spatio-temporal scales, such as modeling the temperature distribution over the globe for the next decade or resolving the finer vortices and shocks often encountered in hypersonic flow problems, both critical problems at the present time. 
The recent innovations in the field of ML primarily originate from the domains of natural language processing \cite{hinton2012deep} and computer vision \cite{krizhevsky2012imagenet}. Our work exploits and further develops the main idea of a recently proposed method called \textit{diffusion models} \cite{ho2020denoising} (used in generative AI)  for solving forward partial differential equations (PDEs).

Solving time-dependent PDEs is an essential topic for the scientific community. This centuries-old research involves 1) formulating a problem from physical domains, biological research, chemical reactions, etc., and 2) solving them using discretization-based approaches like finite-element \cite{hughes2012finite}, finite-difference \cite{godunov1959finite}, finite-volume \cite{eymard2000finite}, or spectral methods \cite{karniadakis2005spectral}. 
These numerical solvers have widespread application across diverse scientific domains, such as wave propagation \cite{kundu2014acoustic,ovadia2021beyond}, computational mechanics \cite{cai2021physics,zhang2022hybrid}, materials science \cite{dingreville2020benchmark, oommen2022learning}, fluid dynamics \cite{patera1984spectral,kim1987turbulence}, seismic imaging \cite{sleeman1999robust}, etc. 
However, if the underlying mathematical operators that govern the temporal evolution of the system are non-linear and/or there are observational data available, the task of assimilating and simulating such processes using discretization-based numerical methods can become increasingly challenging and computationally expensive. 
The burden associated with the traditional numerical solvers is further increased when separate simulation runs become mandatory for every new initial condition. Moreover, in certain application domains, such as autonomy and navigation or robotics, real-time inference is required.
SciML methods such as neural operators specifically address these issues by significantly reducing the associated computational costs \cite{lu2021learning, li2020fourier, tripura2023wavelet, cao2023lno}.

Several methods for solving PDE-related problems using ML methods, and specifically transformers, have been proposed \cite{RAISSI2019686,lu2021learning,ovadia2023vito}. 
Herein, we solve the forward problem of modeling a time-dependent PDE by training the neural operator to accurately estimate the state field variable at a later time from a given initial condition. 
Specifically, forecasting a continuous in-time solution in real-time for a plurality of initial conditions and extrapolating beyond the training domain is the main focus of this work.
Solving PDE-related problems involves several challenges. 
Two main challenges are generalizations for different problem conditions and dependence on the physical domain's discretization. 
To tackle the first, we utilize tools from the growing field of operator learning \cite{li2020fourier,lu2021learning}, where we use learning techniques to map a function space to another one. 
Thus, we are able to learn a family of solutions of PDEs corresponding to a family of initial conditions. 
For the second challenge, we propose a method that, while being dependent on the spatial discretization, is continuous in the temporal evolution of the solution, which is a prominent challenge in solving dynamical systems.

Recent works \cite{rahman2022u, takamoto2022pdebench} have demonstrated the efficiency of U-Net based architectures for modeling time-dependent PDEs. However, the outputs of these U-Net-based architectures are discrete in time. Gupta \emph{et al.} \cite{gupta2022towards} performed a systematic ablation study to analyze the significance of Fourier layers, attention mechanism, and parameter conditioning in a U-Net-based neural operator.  
The DiTTO method proposed here is a diffusion-inspired model \cite{sohl2015deep}. The common use of diffusion models involves a generative process used to create data samples. It incorporates a Markov chain of diffusion steps, where in each stage a different texture is added to the data sample. Usually, the texture is noise, so new noise distributions are incrementally added in each step. The models have also been used with other kinds of textures, for example creating cartoonish images from plain ones. Herein, we use a similar framework, but instead of conditioning on the noise distribution, we do so for the {\em temporal evolution.} We explore several implementations and training strategies, in addition to the diffusion models themselves. These enhanced methods form the class of explored DiTTO models. Importantly, we demonstrate how this framework can be used for extrapolation, i.e., it can make accurate predictions for samples outside the time interval it was trained to handle.



\section*{Results}
\label{sec:results}


We approximate the time evolution of a PDE solution (forward process). Instead of incrementally adding noise to the inputs, as done with diffusion models, we incrementally evolve the PDE solution over time. We replace the noise level parameter $\varepsilon$ with the temporal variable $t$. Then, we use the conditioning capabilities of diffusion models to learn the relations between the initial condition, the PDE solution, and the time domain. After the training is complete, the model can interpolate between the initial and final time, creating a numerical solution that is continuous in time. We define its time evolution $\{x_t]\}$ as the following process:
\begin{equation}\label{continuous_process}
    \{x_t | \; t \in [0,t_{final}], \; x_t:= u(\textbf{x}, t) \},
\end{equation}
where $u$ is the solution of the differential equation we attempt to solve. Using this notation, the operator learning problem becomes:
\begin{equation}
        x_0 \longrightarrow x_t, \: \forall t \in {[0, t_{final}]},
\end{equation}
\begin{equation}\label{eq:ditto_operaor_mapping}
        x_t \approx \mathcal{G}(x_0)(t), \: \forall t \in {[0, t_{final}]}, 
\end{equation}
where $\mathcal{G}$ represents the surrogate operator, DiTTO, where the operator learning technique we employ is the diffusion process. It is discrete, while \eqref{continuous_process} is continuous. We discretize $\{x_t\}$ by taking a partitioning $\{t_n\}_{n=0}^T$ of $[0, t_{final}]$, where $0 = t_0 < t_1 < \ldots < t_{T-1} < t_T = t_{final}$. The discrete process is then defined as $\{x_n\}_{n=0}^T$, where $x_n := u(\textbf{x}, t_n)$. In PDE terms, given an initial condition $x_0$, we approximate the analytic solution at a set of specific future time steps $\{t_n\}_{n=1}^T$. In operator learning terms, we map a family of functions of the form $x_0=u(\textbf{x}, 0)$ to another family of functions of the form $u(\textbf{x}, t)$. 

As outlined before, the role of the neural network in diffusion models is to perform conditional denoising in each step. We repurpose this network structure to solve a PDE-related problem. Since $x_0, x_1, \ldots, x_T$ are directly taken from the analytical solution, we have no noise in this process. Therefore, there is no need for denoising. Thus, we replace the conditional denoising operation with a conditional temporal evolution  $(x_0, t_n) \longrightarrow x_n$.


Next, we describe the DiTTO architecture. 
The network receives two main inputs: the initial condition $x_0 = u(\textbf{x},0)$ and a time $t=t_n$. Recall that $x_0$ is a $d$-dimensional tensor, and $t$ is a nonnegative scalar. For the temporal input $t$, we use an embedding mechanism based on the original Transformer positional encoding \cite{vaswani2017attention}. Each scalar $t$ is mapped into a vector of size $d_{emb}$, and then passed through a simple multi-layer perceptron (MLP) with two linear layers and a GELU \cite{hendrycks2016gaussian} activation function. 

\begin{figure}
    \centering
    \includegraphics[scale=0.5]{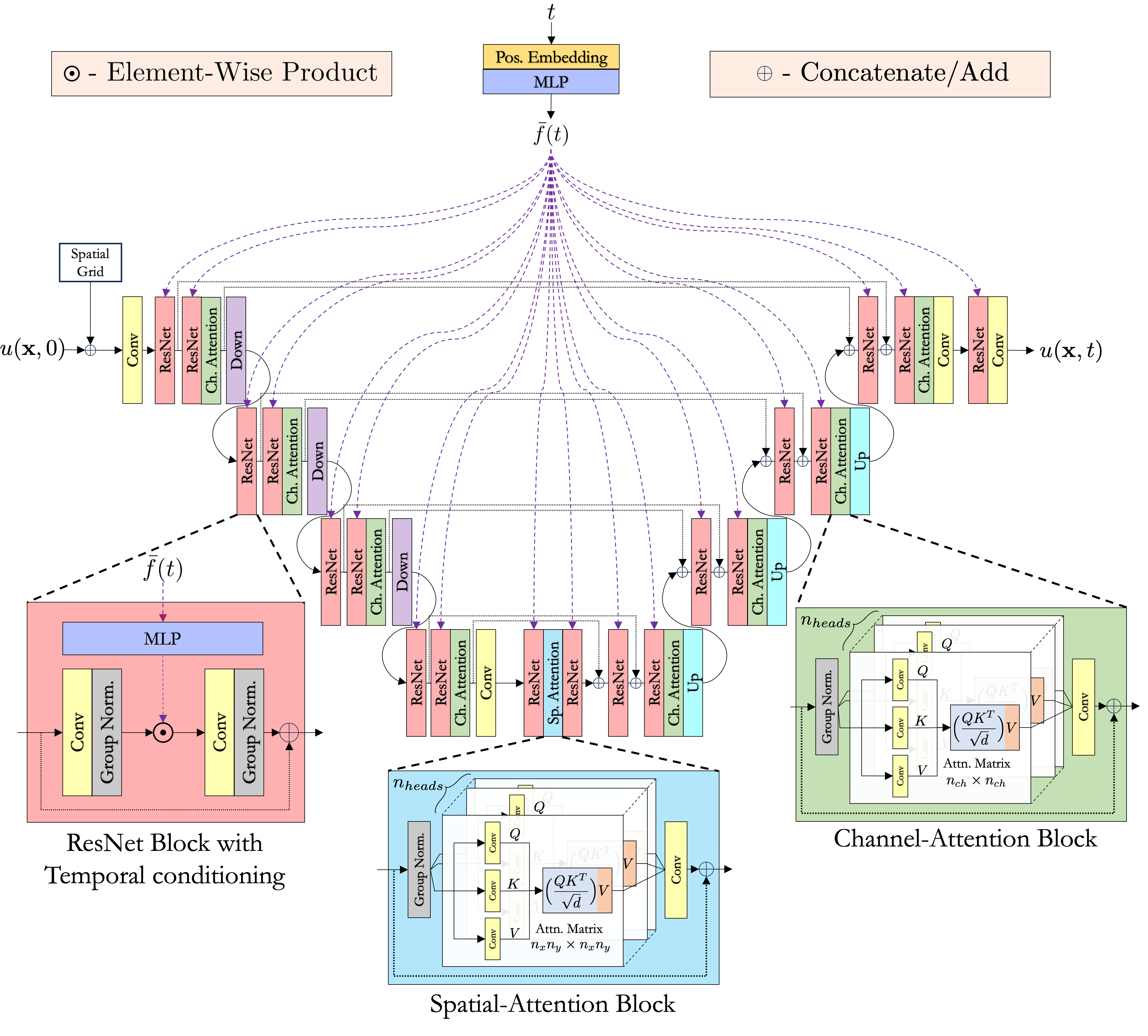}
    \caption{DiTTO architecture. The discretized initial condition $u(\textbf{x}, 0)$ concatenated with the corresponding spatial grid, and the desired time $t \in \mathbb{R}^+$ are the respective inputs to the U-Net and the time-embedding network comprising DiTTO. The U-Net illustrated here consists of ResNet blocks with temporal conditioning, a Spatial-Attention block, and Channel-Attention blocks at 4 levels of coarseness, and the corresponding residual connections across the same levels. The ResNet block conditions the non-linear representations of $u(\textbf{x},0)$ with respect to the temporal embedding vector, $\vec{f}(t)$, by performing element-wise multiplication across the channels. Spatial-Attention and Channel-Attention blocks learn to extract correlations across space and channels respectively.}
    \label{fig:ditto_architecture}
\end{figure}

For the spatial input $x_0$, we concatenate it with a discrete spatial grid and provide it as an input to a U-Net \cite{ronneberger2015u}. We use a U-Net variant common in many diffusion models, such as DDPM \cite{ho2020denoising}. It follows the backbone of PixelCNN++ \cite{salimans2017pixelcnn}, a U-Net based on a Wide ResNet \cite{zagoruyko2016wide, he2016deep}. A sketch of the spatio-temporal architecture is given in \Cref{fig:ditto_architecture}, and more details can be found in \nameref{sec:methods}.
This architecture is not limited to a specific dimension. The same mechanism can be implemented for $d$-dimensional problems, where $d \in \{1, 2, 3 \}$. The only major difference is the usage of $d$-dimensional convolutions for the relevant problem. 

Furthermore, we extend DiTTO to develop three variants: DiTTO-s, DiTTO-point and DiTTO-gate. DiTTO-point is a memory-efficient version and DiTTO-gate incorporates a gated sub-architecture motivated by Runge-Kutta methods. We also demonstrate the efficacy of two training strategies - 1) randomly sub-sampling the trajectory timesteps considered at each epoch (DiTTO-s), and 2) adopting the temporal-bundling \cite{brandstetter2022message} for enhancing the forecast capabilities. 

\begin{figure}
    \centering
    \includegraphics[scale=0.5]{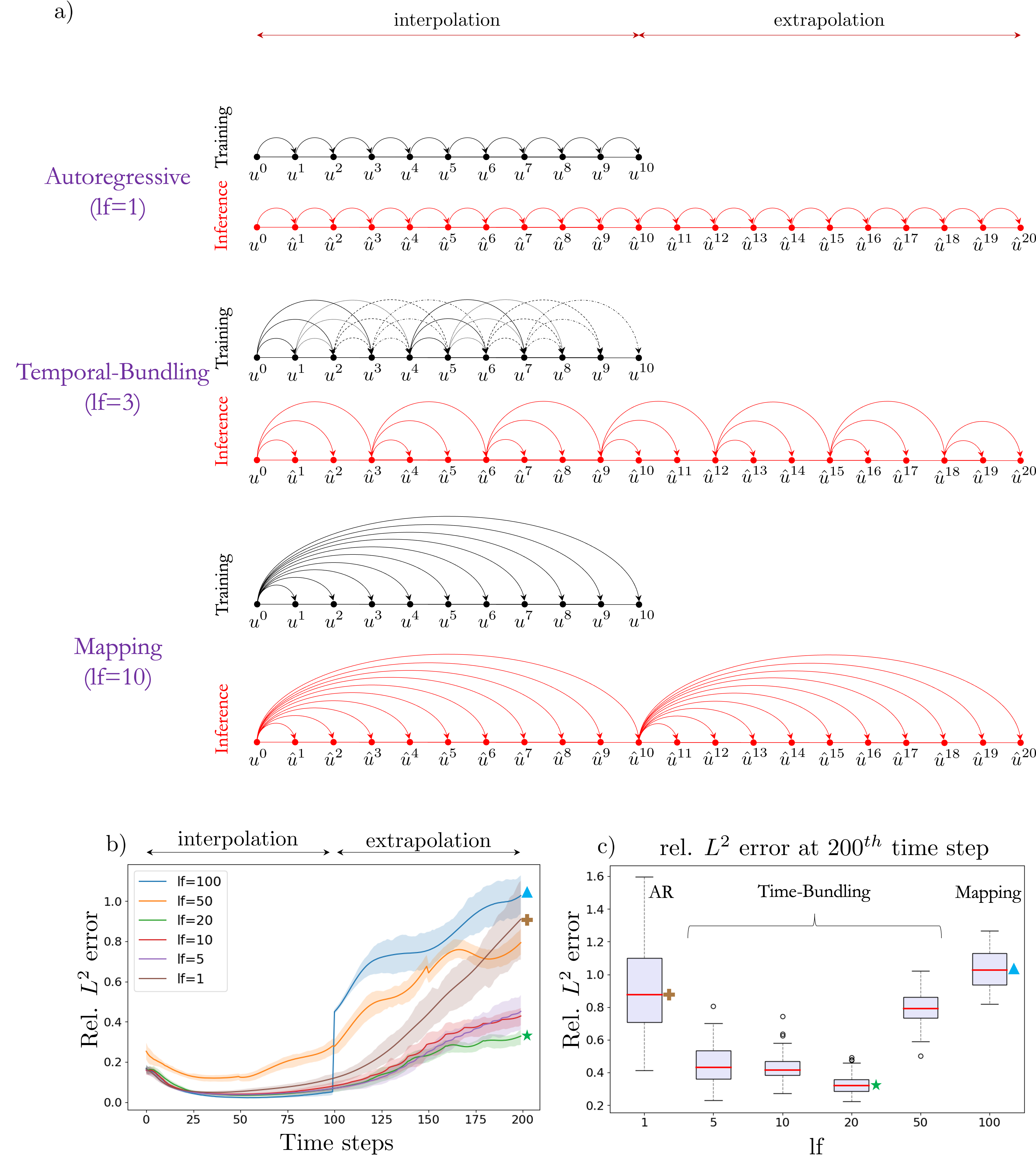}
    \caption{Temporal-bundling for efficient extrapolation. a) demonstrates 3 types of time-series modeling strategies for extrapolating beyond the training interval. For display purposes, we consider a time-series with 20 time steps. b) visualizes the error accumulation for DiTTO models with different look-forward windows ($lf$) trained in the time interval 0-100 and further extrapolated from 100-200. c) shows the relative $L^2$ errors (in percentage) at the 200$^{th}$ time step, corresponding to DiTTO models with different lf. We note that the results in b) and c) come from the test dataset with previously unseen 100 initial conditions. The symbols in (b-c) denote the corresponding final time.}
    \label{fig:extrapolation}
\end{figure}

Next, we investigate the capabilities and limitations of the proposed DiTTO method on a wide variety of time-dependent non-linear systems in multiple dimensions.
First, we discuss the temporal-bundling training technique and illustrate accurate extrapolation on the climate problem. 
Next, we demonstrate the efficiency of DiTTO in approximating operators with high spatial gradients by learning high Mach number hypersonic flow past a double-cone geometry. 
We also train DiTTO to learn the following benchmark PDEs: 1D Burgers' equation, 2D incompressible Navier-Stokes with high and low Reynold's numbers, and 2D and 3D acoustic wave equation problems. 
We compare the DiTTO and its variants: DiTTO-s that utilizes fast sampling, DiTTO-point that reduces high-dimensional problems to 1D, and DiTTO-gate  that incorporates a novel convolutional block inspired by stencil-based numerical methods. 
We compare the proposed methods to the FNO method \cite{li2020fourier}, as well as a standard U-Net model with attention. We use the U-Net architecture, except we remove all temporal conditioning. We compute the relative $\textit{L}^2$ error for each method and compare it to the ground truth data. 

For each problem we either synthetically create a dataset for training or use data from external sources. Details regarding the data collection process and the solvers used for each problem are presented in the corresponding sections. In each case, the spatiotemporal resolution of the numerical solution is kept fixed. The size of the spatial grid, denoted by $N_x, N_y$, $N_z$, is determined by the dimensionality of the problem. We use the same spatial grid for training and inference. For the number of timesteps $N_t$ (following standard PDE notation), we use different resolutions for training and inference to test the temporal interpolation and super-resolution capabilities of DiTTO. We train all models with, for example, $N_t^{train} = 50$ time steps, and test on $N_t^{test} \in \{10, 20, 50, 100, 200 \}$ (exact details are given per problem). These choices of $N_t^{test}$ allow us to examine the results in three different regimes. First, when $N_t^{test} = 50$, we test the model on the same temporal resolution it was trained on. We note that the model is able to handle coarser temporal grids for $N_t^{test} \in \{10, 20 \}$. Finally, we consider a zero-shot super-resolution scenario in time for $N_t^{test} \in \{100, 200\}$, exploring the interpolation capabilities of the model on unseen temporal discretizations. Note that DiTTO and its variants are entirely meshfree in time due to the time conditioning mechanism.



\subsection*{Extrapolation in time}


In our application, extrapolation is a challenging problem due to the inherent nature of the data-driven surrogate networks to overfit the training distribution. 
Next, we discuss a training strategy that partially alleviates the difficulties associated with extrapolation in time in using DiTTO.
As an example, we train DiTTO on the incompressible Navier-Stokes equations  with $Re=20$ for investigating the ability to extrapolate in time. The dataset consists of 1000 trajectories with 200 time steps. It was randomly split 
into training, validation and testing datasets in the ratio 80:10:10. 
%
During the training of DiTTO, the model is exposed only to the first 100 time steps. We train DiTTO with three types of time-series modeling strategies demonstrated in part a) of \Cref{fig:extrapolation} - i) autoregressive (look-forward window $lf=1$), ii) temporal-bundling ($1<lf<nt$) \cite{brandstetter2022message}, and iii) mapping ($lf=nt$). We analyze their ability to extrapolate beyond the 100$^{th}$ time step. Specifically, we train 6 different DiTTO models with $lf$=1, 5, 10, 20, 50, 100($=nt$). During the training, DiTTO learns a mapping $u(\bar{x},t)\rightarrow u(\bar{x}, t+lf)$ such that $t+lf$ is less than the number of time steps, $nt=100$. Hence, each trajectory in the training dataset is split into $nt-lf+1$ sub-trajectories. During the inference stage, DiTTO leaps from $\hat{u}(\bar{x},t)$ to $ \hat{u}(\bar{x}, t+lf)$. Because the ground truth is not available, except at $t=0$, we consider $u$ as the ground truth and $\hat{u}$ as a prediction from DiTTO. 
%
In \Cref{fig:extrapolation}(b,c) we observe the minimum error with the lowest uncertainty occurs at $lf=20$, suggesting that the temporal-bundling technique \cite{brandstetter2022message} offers a sweet spot between the autoregressive and the mapping strategies for extrapolating in time, with lower rates of error accumulation. 
Next, we investigate the extrapolation abilities of the temporal-bundling strategy for the climate problem.


\subsection*{Climate Modeling}

Climate models are complex, involving non-linear dynamics 
and multi-scale interactions between multiple variables. Describing climate behavior using PDEs is not a straightforward task. Consequently, there are many different numerical models and ML-based models \cite{hurrell2013community, bora2023learning, kissas2022learning, pathak2022fourcastnet, lam2022graphcast, nguyen2023climax} that have been developed to model climate. Furthermore, climate-related problems are challenging to approximate accurately at a low computational cost. Our goal is to efficiently learn the temporal evolution of the surface air temperature across the globe, and make accurate forecasts by extrapolating beyond the training period. 

\begin{figure}
    \centering
    \includegraphics[scale=0.5]{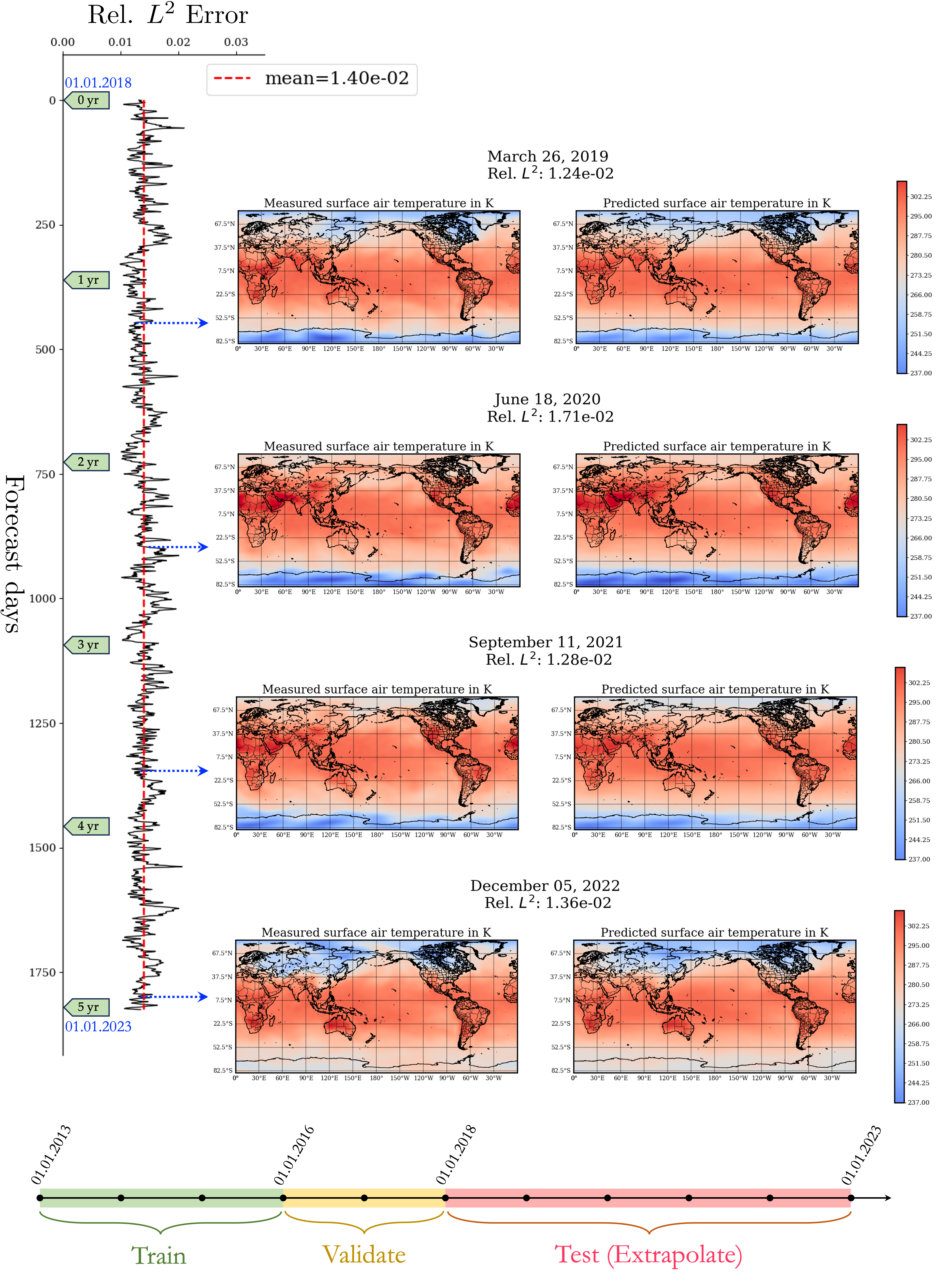}
    \caption{Five-year climate forecast (extrapolation). DiTTO is trained on the global temperature data available in \href{https://psl.noaa.gov/data/gridded/index.html}{NCEP-NCAR Reanalysis 1 dataset} from years 2013 to 2015, validated from 2016 to 2017, and extrapolated from 2018 to 2022. The contour plots on the left and right columns correspond to snapshots from five years of measured and forecasted global temperature distribution respectively. The four rows correspond to the global temperature profiles in Spring, Summer, Autumn, and Winter. The effectiveness of the proposed approach to learn the seasonal climate trends is visually illustrated using the contour plots and quantitatively reported on the error accumulation curve on the left: mean error is 1.4\%.}
    \label{fig:climate}
\end{figure}

To achieve this goal, we use a publicly available climate dataset provided by the Physical Sciences Laboratory meteorological data: \url{https://psl.noaa.gov/data/gridded/index.html} \cite{kalnay2018ncep}. 
This data includes measurements of climate parameters over time, projected onto a spatial grid. 
Specifically, we use the daily average surface temperature data (at 1000 millibar Pressure level) from January 1, 2013, to December 31, 2015, for training; January 1, 2016, to December 31, 2017, for validating; and January 1, 2018 to December 31, 2022 for testing (see \Cref{fig:climate}). 
The data is projected to a spatial grid of dimensions $144 \times 72$, which corresponds to a resolution of ${2.5}^{\circ}$ in both latitude and longitude. 
For this particular test case, we split the data to train, validation and test sets in the ratio 30:20:50, for investigating the real-time long-term temporal extrapolation capability of DiTTO. 

The experiments demonstrated in \Cref{fig:extrapolation} motivated the adoption of the temporal-bundling technique with $lf=365$ for training DiTTO. 
The training dataset can be interpreted as a single trajectory consisting of 1095 global temperature snapshots arising from the initial condition on Jan. 1, 2013. 
We utilize temporal-bundling with $lf=365$ to split this trajectory into 730 sub-trajectories, each comprising 365 time steps. 
We further reduce the computational costs associated with the training and inference, by performing a proper orthogonal decomposition and training DiTTO to learn the evolution of the eigen coefficients corresponding to the five dominant eigenvectors estimated from the global temperature profiles in the training dataset.  
%
The results in \Cref{fig:climate} indicate that DiTTO is able to extrapolate for long time intervals without any substantial accumulation of errors.  The contour plots illustrate the measured (left) and predicted (right) global temperature profiles in Spring, Summer, Autumn and Winter. DiTTO is able to reduce the growth of the relative $L^2$ error across the 5 years of extrapolation with a mean of 0.014.  





\subsection*{Hypersonic flow}
\label{subsubsec:hypersonics}

Next, we train DiTTO to learn the inviscid airflow around a double-cone object flying at a high Mach number.  The flow physics at hypersonic Mach numbers around the double-cone geometry features complex transient events,  stationary bow shock at the leading nose of the cone, and the interaction of the oblique shock wave originating from the leading edge with the bow shock formed around the upper part of the geometry of the cone. The narrow band between the wall of the double-cone geometry and the bow shock wave is called the shock layer. In the shock layer, due to the interaction of shock waves, a triple point forms and generates vortical flow structures that move downstream. This interaction is challenging to capture using numerical solvers. The details regarding data creation and the governing equations are given in SM. 
We now present two experiments conducted for this hypersonic problem.

\begin{figure}[!ht]
    \centering
    \includegraphics[width=1\linewidth]{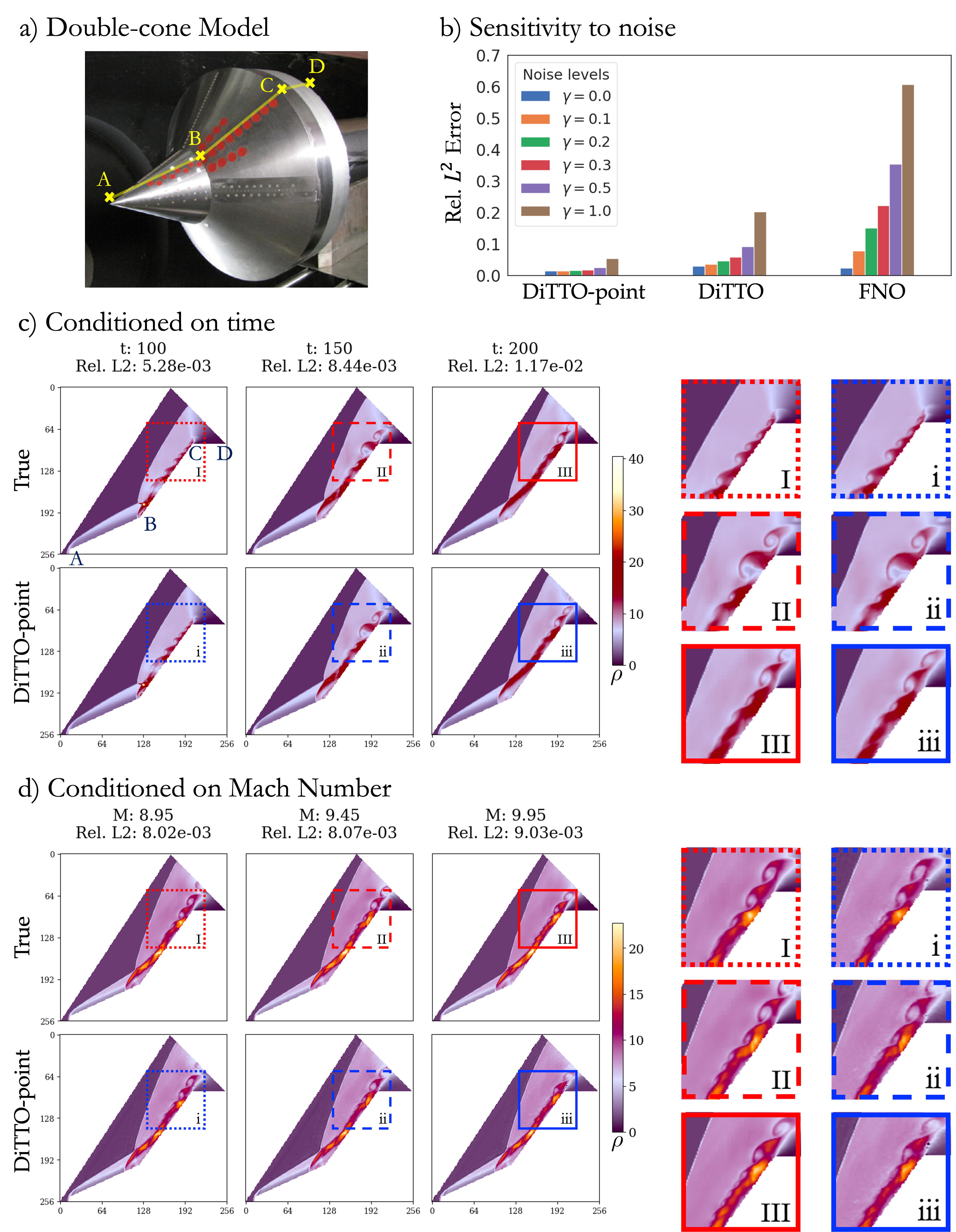}
    \caption{Modeling high-speed flow past a double-cone. 
    a) The image of the double-cone model installed in the LENS XX tunnel \cite{rouhi2017assessment}. 
    b) The histogram illustrates the sensitivity of (i) DiTTO-point, (ii) DiTTO, and (iii) FNO to different noise levels $\gamma$.
    c) Temporal evolution of the density field ($\rho$) corresponding to test Mach number $(M)=9.18$ around the double cone geometry predicted by DiTTO-point and its comparison with the solver simulation.
    d) Additional illustration of DiTTO-point conditioned on $M$, and trained to predict the final state of $\rho$. A video illustrating the flow is included in SI.} 
    \label{fig:hypersonics}
\end{figure}

First, we learn the temporal evolution of the density field near the double cone geometry. Specifically, we train the neural operator to learn the mapping from the incoming horizontal velocity field to the time-dependent density field around the double cone structure. The dataset comprises only 61 trajectories corresponding to Mach numbers $M \in [8,10]$. It is split into training, validation, and testing datasets in the ratio 80:10:10. We use a cosine annealing type learning rate scheduler starting from $10^{-3}$ that decays to 0 during the course of the training. In this study, we compare FNO, U-Net, DiTTO and DiTTO-s. Complete results are reported in SM. 
We observe in \Cref{fig:hypersonics} that DiTTO is able to accurately resolve the vortices close to the surface of the double cone. Quantitative results are shown in \Cref{tab:full_comparison}.


Additionally, we perform an experiment, where instead of conditioning on time as in most of the examples presented, we explore the ability to condition DiTTO on a different quantity: the Mach Number. We train the surrogate operator to learn the mapping from the density field at a specific time step and $M=8$ to the density field at the same timestep but at different Mach numbers in the range $M \in [8,10]$. We compare DiTTO with other neural operators. Our results suggest that DiTTO serves as an accurate surrogate conditioned on any scalar parameter, and not necessarily time. An illustrative result is presented in \Cref{fig:hypersonics}. 

We also evaluate how well the DiTTO method can handle noise in the hypersonic scenario. We use zero-mean Gaussian additive noise, which is a common choice. Since different PDEs can behave quite differently, we ensure that the variance of the Gaussian noise is dependent on the input data. 
The operation of adding noise is given by:
$    x \longmapsto x + \gamma \mathcal{N}(0, \sigma^2_{\mathrm{D}}),$ 
where $x$ is an input sample in the dataset $\mathrm{D}$, $\sigma^2_{\mathrm{D}}$ is the variance of the entire dataset, and $\gamma$ is the desired noise level, e.g., $\gamma = 0.1$ is equivalent to $10 \%$ noise. Importantly, we add noise only to the testing set, hence, the model does not encounter noisy samples during the training process. 
Results are shown in \Cref{tab:noise} and \Cref{fig:hypersonics}. For the case without noise, we see that all the models exhibit low errors, with the DiTTO-point model having the lowest error, followed by FNO. However, when adding noise,  both the DiTTO and DiTTO-point methods are able to maintain lower errors compared to FNO. Furthermore, DiTTO-point is able to handle very large amounts of noise - up to a $100 \%$ with respect to the standard deviation - and still maintain a low error rate, thus demonstrating excellent noise robustness.

\subsection*{2D and 3D Wave Propagation}

Next, we consider the following formulation of the acoustic wave equation in 2D and 3D \cite{evans2022partial, jost2012partial}:

\begin{equation}\label{eq:gen_wave_eq}
	\begin{cases}
		u_{tt}(\textbf{x}, t)=c^2(\textbf{x}) \Delta u(\textbf{x}, t) & \textbf{x} \in \Omega, 0\leq t \leq 2, \\
		u(\textbf{x}, 0)=u_0(\textbf{x}) & \textbf{x} \in \Omega, \\
		u_t(\textbf{x}, 0)=0 & \textbf{x} \in \Omega,
	\end{cases}
\end{equation}

\noindent where $u(\textbf{x},t)$ is the wave amplitude or acoustic pressure, $c(\textbf{x})$ is the wave propagation speed, $\Omega = \left[0, \pi\right]^d$ where $d \in \{2, 3\}$ is the physical domain, $\Delta$ is the 2D/3D Laplacian operator, and the PDE is subject to homogenous Dirichlet (fully reflective) boundary conditions. We set the speed of sound of the medium as $c(x,y) = (1 + \sin{(x)} \sin{(y)})$ in 2D and $c(x,y,z) = (1 + \sin{(2x)} \sin{(y)} \sin{(z)})$ in 3D.

The initial condition $u_0$ is chosen to be a Gaussian source of the form:
\%begin{equation}\label{eq:g_erup}
 $   u(\textbf{x}, 0) = e^{-\left(\frac{||\textbf{x}-\textbf{x}_c||_2^2}{10}\right)}.$ 
%
To create the dataset, we generate several initial conditions of the same type, randomly varying the spatial location $\textbf{x}_c$ of the center of the source. The locations are sampled using a discrete random uniform distribution on the indices of the grid. We generate the numerical solutions using a finite-difference numerical scheme with a grid of $N_x = N_y = 64$ spatial nodes in the 2D case, and $N_x = N_y = N_z = 32$ in the 3D  case.

\begin{figure}[!ht]
    \centering
    \includegraphics[scale=0.6]{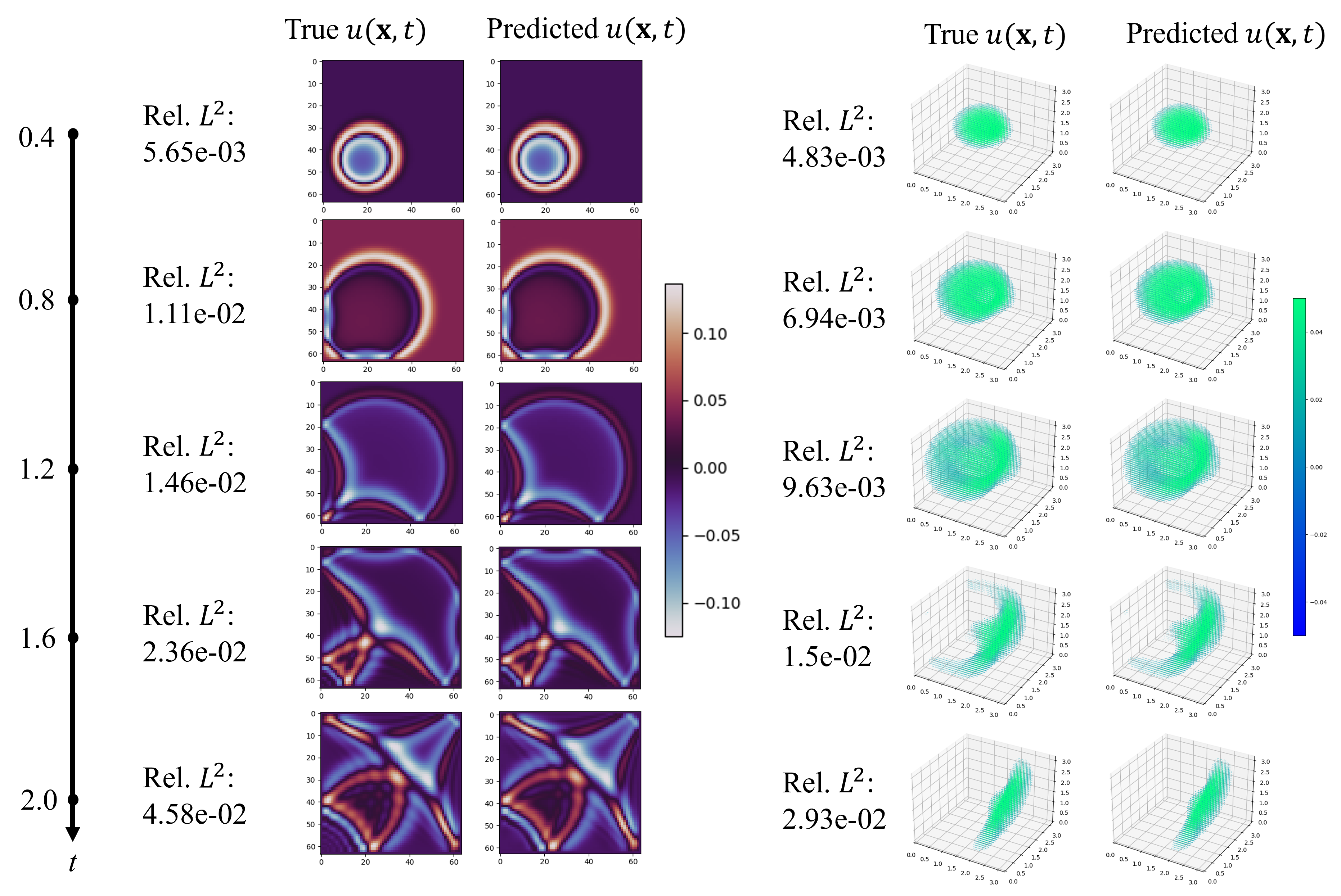}
    \caption{Illustrations of 2D (left) and 3D (right) wave propagation numerical experiments. The predictions are made using the DiTTO-s variant. The initial emmitted wave is a small Gaussian eruption bell in 2D and sphere in 3D. The boundaries are reflective. Videos illustrating the process are included in SI.}
    \label{fig:wave_eq_3d}
\end{figure}

The results are shown in \Cref{tab:full_comparison}. DiTTO and DiTTO-s both achieve the lowest errors. We note that the FNO and the U-Net fare much worse compared to the other PDE cases when changing the discretization. Specifically, the errors of DiTTO and DiTTO-s are an order of magnitude lower than FNO. We hypothesize that this might be related to the use of Dirichlet boundary conditions instead of periodic ones. Another reason might be the sparsity of the data in this case, which causes the solution to change rapidly over time and have sharp features (see \Cref{fig:wave_eq_3d}).

\begin{table}[!htb]
\centering 
\begin{tabular}{|ll|ccccc|}
\hline 
 Scenario & Model & $N_t^{test} = 10$ & $N_t^{test} = 20$ & $N_t^{test} = 50$ & $N_t^{test} = 100$ & $N_t^{test} = 200$ \\
\hline\hline
\multirow{4}{9em}{2D time-dependent hypersonic flow} 
& DiTTO               &             0.2253 &             0.1264 &             \textbf{0.0224} &              0.0601 &              0.0815 \\
& DiTTO-point         &             0.2263 &             0.1280 &             0.0251 &              \textbf{0.0573} &              \textbf{0.0779} \\
& DiTTO-s             &             0.2243 &             0.1321 &             0.0655 &              0.0721 &              0.0854 \\
& DiTTO-point-s       &             \textbf{0.2089} &             0.1277 &             0.1175 &              0.1364 &              0.1474 \\
& DiTTO-point-s-gate  &             0.2352 &             0.1490 &             0.0722 &              0.0699 &              0.0790 \\
& FNO                 &             0.4702 &             \textbf{0.1212} &             0.0510 &              0.0746  &            N/A \\
\hline
\multirow{4}{9em}{2D Navier Stokes \\ $Re \approx 2,000,$ \\ $\#samples=5,000$} 
& DiTTO-s             &    \textbf{0.1701} &    \textbf{0.1113} &    \textbf{0.0919} &     \textbf{0.0913} &              0.0923 \\
& DiTTO-point-s-gate  &             0.1936 &             0.1322 &             0.1014 &              0.0944 &     \textbf{0.0919} \\
& DiTTO-point-s       &             0.1721 &             0.1132 &             0.0945 &              0.0942 &              0.0953 \\
& FNO                 &             0.3923 &             0.1986 &             0.1260 &              0.1312 &              0.1398 \\
\hline
\multirow{4}{9em}{2D wave equation} 
& DiTTO               &    \textbf{0.0091} &             0.0246 &    \textbf{0.0077} &              0.0379 &              0.0481 \\
& DiTTO-point         &             0.0140 &    \textbf{0.0137} &             0.0122 &     \textbf{0.0324} &              0.0391 \\
& DiTTO-s             &             0.0343 &             0.0320 &             0.0301 &              0.0354 &     \textbf{0.0376} \\
& DiTTO-point-s       &             0.0519 &             0.0486 &             0.0450 &              0.0540 &              0.0576 \\
& DiTTO-point-s-gate  &             0.0335 &             0.0316 &             0.0296 &              0.0382 &              0.0412 \\
& FNO                 &             1.3253 &             0.4903 &             0.1460 &              0.2093 &              0.2636 \\
& UNet                &             1.3956 &             1.3998 &             0.0202 &              1.2501 &              1.2595 \\
\hline
\multirow{4}{9em}{3D wave equation} 
& DiTTO-s             &    \textbf{0.0192} &    \textbf{0.0166} &    \textbf{0.0151} &     \textbf{0.0148} &     \textbf{0.0147} \\
& DiTTO-point-s       &             0.0384 &             0.0350 &             0.0331 &              0.0326 &              0.0324 \\
& DiTTO-point-s-gate  &             0.0373 &             0.0339 &             0.0319 &              0.0314 &              0.0311 \\
\bottomrule
\end{tabular}
\caption{A full error analysis of all cases comparing the various DiTTO methods with FNO and U-Net across different temporal resolutions. Results for additional cases are shown in SI.}
\label{tab:full_comparison}
\end{table}

\section*{Discussion}
\label{sec:discussion}


We have developed a new neural operator, DiTTO, for modeling time-dependent PDEs. 
DiTTO comprises of U-Net with spatial and channel-wise attention blocks and modified ResNet blocks that incorporate a temporal conditioning operation. 
The attention blocks allow DiTTO to extract spatial correlations as well as the correlations between the different representations of the inputted field variable. 
The ResNet block with the temporal conditioning operation learns to project the several spatial representations of the inputted field variable with respect to coefficients which are functions of time. 
These coefficients are linear projections of a collection of functions of time learned by an MLP. Unlike FNO and U-Net, which predict the time-evolving systems at certain discrete timesteps, DiTTO estimates the underlying dynamical operator in a continuous sense. 


The temporal conditioning operation embedded into DiTTO enables it to estimate the state of the field variables at any desired time step within the training time interval. This makes DiTTO an ideal neural operator for temporal super-resolution tasks. 
We validated our hypothesis through extensive computational experiments. 
Specifically, we trained the neural operators - DiTTO (and its variants), FNO and U-Net, on the training datasets with a temporal resolution of $N_t^{train}=50$ and made inferences on the test datasets with temporal resolutions of $N_t^{test} = 10,20,50,100,200$.  
\Cref{tab:full_comparison} indicates that DiTTO and its variants outperform the other neural operators for $N_t^{test}<50$ for all the benchmarks considered, except for one case the Navier-Stokes 2d.
For $N_t^{test}>50$, DiTTO and its variants outperform the other neural operators considered here for all the test cases in this study. 
Therefore, our experiments demonstrate that DiTTO offers the best zero-shot temporal super-resolution with the least rate of increase in error for $N_t^{test}>N_t^{train}$ compared to the other neural operators considered here.

Neural operators learn time-dependent systems from the mappings between the initial condition and the state of the field variable $u$ at $t^{th}$ time step, where, $0<t<T$. 
However, a neural operator trained only on $T$ mappings per trajectory would fail miserably if it infers beyond the $T^{th}$ timestep. To extrapolate beyond the $T^{th}$ timestep, the inferred output $\hat{u}$ at the $T^{th}$ time step must be fed as an input to the neural operator.
Subsequently, the neural operator collapses because the $\hat{u}(T)$ is very dissimilar to any of the inputs ($u(0)$) the neural operator ever encountered during its training. 
On the other hand, training the neural operator autoregressively, from $t^{th}$ to the $t+1^{th}$ timestep may not be a suitable choice because the error has already accumulated $T$ times by the time DiTTO predicts the $T^{th}$ timestep. 
To this end, we adopted the temporal-bundling technique 
and demonstrated its efficiency compared to the other two training strategies in \Cref{fig:extrapolation} b) and c). 
The temporal-bundling offers a lower error with lower uncertainty for temporal extrapolation tasks. The three training strategies discussed in this work are independent of the neural operator and hence they can be combined with other operators as well.

Lastly, we observed that DiTTO scales in dimension well compared to its competitors. We also experimented with training DeepONet. For time-dependent 1D cases, it required a very large amount of examples to learn both different initial conditions and the entire time-dependent process before converging. In multidimensions, the required amount of data was too much for our hardware to handle, and we eventually removed the DeepONet examples. Simple UNet examples were added to the comparisons with DiTTO, but could not capture complex solutions, e.g., it was not possible to train the U-Net on 3D time-dependent problems, due to their large memory footprint. For the FNO method,  the issue arises since adding a temporal element is done by adding a dimension, so in the view of the U-Net and FNO, our 3D time-dependent problems are actually four-dimensional, which introduces excessive computational cost.
DiTTO, on the other hand, was able to scale well since the temporal axis is taken care of by conditioning. Also, with DiTTO-point, we reduced each problem to a 1D problem, reducing the computational cost for high-dimensional problems. DiTTO-point exhibits comparable error to the original DiTTO architecture, despite being one-dimensional. This is attributed to the positional embeddings in space, which enable the model to keep track of the original spatial locations.


\section*{Methods}
\label{sec:methods}



\subsection*{DiTTO - Diffusion-inspired Temporal Transformer Operator}

DiTTO is a multiscale neural operator architecture that can be trained to effectively approximate time-dependent PDEs within a data-driven framework. 
The input to DiTTO is a vectorized representation of the initial condition ($u(\textbf{x}, 0)$) at predefined spatial grid locations and the query time ($t$). 
DiTTO is trained to output the vectorized representation of $u(\textbf{x},t)$, making the prediction of DiTTO discrete in space and continuous in time. 
DiTTO is built with two components: (i) a U-shaped convolutional network (U-Net), and (ii) a time-embedding network, as illustrated in \Cref{fig:ditto_architecture}. 

\textbf{U-Net with attention blocks:} The U-Net type architecture enables DiTTO to learn the underlying dynamics at different latent scales. 
The spatial and channel attention blocks within the U-Net facilitate efficient extraction of spatio-temporal correlations.
Specifically, we implement the scaled dot product attention by linearly projecting the input representation to query ($Q$), key ($K$) and value ($V$) vectors using separate 1$\times$1 convolution layers.  
Next, we compute the attention matrix and calculate the output as,
\begin{equation}
    \text{out} = \frac{QK^T}{\sqrt{d}}V,
\end{equation}
where $\frac{QK^T}{\sqrt{d}}$ is the attention matrix with a shape of $n_{channels}\times n_{channels}$ or $n_{pixels} \times n_{pixels}$ for channel or spatial attention blocks respectively.

\textbf{Temporal-conditioning:} The time-embedding network conditions the U-Net with respect to the scalar parameter $t$ through element-wise product operation across the channels making the DiTTO output continuous in time. 
The temporal conditioning mechanism is incorporated into the ResNet block, which is another building block of the U-Net. 
See Supplementary Information (SI) for more details on the variants of DiTTO and diffusion models in general. 

\subsection*{Extrapolation in time}

Neural operators are trained to learn time-dependent systems with the help of datasets containing trajectories arising from multiple initial conditions. 
Conventionally, such neural operators are trained to learn the mappings from the initial condition to all the available later timesteps of the trajectory. 
However, the neural operators quickly overfit the training timesteps making the scope of temporal extrapolation bleak. 
On the other hand, if the neural operator is trained to predict only the next timestep and consequently predict the trajectories in a purely autoregressive fashion, the accumulation of errors at every timestep hinders accurate generalization. 
Therefore, we train the surrogate operator to predict a bundle of future timesteps thereby minimizing the detrimental effects associated with a large number of autoregressive steps. 
The temporal-bundling training technique also exposes the surrogate neural operator to having intermediate states as inputs and therefore regulates itself from overfitting to having only the initial states as the inputs, as mentioned earlier.
The idea was originally introduced in \cite{brandstetter2022message} and we demonstrate its effectiveness for extrapolation in time in a systematic manner through our experiments illustrated in \Cref{fig:extrapolation}.

\subsection*{Error Metrics}
\label{subsec:error_metrics}
In this work, we compute and report the relative $L^2$ error for quantifying the goodness of the predictions presented. The relative $L^2$ error is defined as,
\begin{equation}
    \text{Rel. $L^2$ error}(u(\textbf{x},t)) = \frac{|| u_{\text{true}}(\textbf{x},t) - u_{\text{pred}}(\textbf{x},t) ||_2}{|| u_{\text{true}}(\textbf{x},t) ||_2}
\end{equation}

\section*{Acknowledgement}

This work was supported by the Vannevar Bush Faculty Fellowship award (GEK) from ONR (N00014-22-1-2795).
It is also supported by the U.S. Department of Energy, Advanced Scientific Computing Research program, under the Scalable, Efficient and Accelerated Causal Reasoning Operators, Graphs and Spikes for Earth and Embedded Systems (SEA-CROGS) project, DE-SC0023191.










\bibliographystyle{unsrt}  
\bibliography{references}  

\newpage
\appendixpageoff
\appendixtitleoff
\begin{appendices}
\crefalias{section}{supp}
\crefalias{subsection}{supp}
\crefalias{subsubsection}{supp}

\section*{Supplementary Information}

\section{Detailed methods}
\subsection{Background}
\subsubsection{Operator learning}
\label{supp:operator_learning}

The standard use of ML models for scientific computations involves fitting a function to map numerical inputs to outputs. These inputs are ordinarily coordinates, materials, boundary conditions, etc., and the outputs are usually solutions of forward PDEs. An example is physics-informed neural networks (PINNs) \cite{RAISSI2019686}, which use a deep neural network to solve PDEs by embedding elements from the PDE problem into the loss function. In this way, the network trains on the given data while using prior information about the problem it is solving. One major drawback is that for each problem, one needs to re-train the network, which is computationally expensive. This includes any changes to the parameters defining the problem, such as the inputs mentioned above.

The field of operator learning seeks to overcome this problem. Instead of fitting a function, one fits a mapping between two families of functions. Mathematically, consider a generic family of $d$-dimensional time-dependent PDE problems of the form:

\begin{align}\label{eq:generic_pde}
\begin{cases}
    \mathcal{L}u(\textbf{x}, t) = f(\textbf{x}, t), & \textbf{x} \in D, t \in [0, t_{final}] \\
    \mathcal{B}u(\textbf{x}, t) = g(\textbf{x}, t), & \textbf{x} \in \partial D, t \in [0, t_{final}]  \\
    u(\textbf{x}, 0) = I(\textbf{x}),  & \textbf{x} \in D 
\end{cases},
\end{align}
where the differential operator $\mathcal{L}$ and forcing term $f$ define the PDE, the boundary operator $\mathcal{B}$ and boundary condition $g$ define the solution on the boundary, $t_{final}$ is the final physical time, $I$ is the initial condition, and $D$ is a Euclidean domain in $\mathbb{R}^d$ with boundary $\partial D$. We assume that the problem \eqref{eq:generic_pde} is well-posed \cite{hadamard1902problemes}, so a unique solution exists.

Let $\mathcal{I}$ be a function space containing initial conditions of \eqref{eq:generic_pde}. Then there exists another space $\mathcal{U}$ that contains their respective solutions. We define an operator $\mathcal{G}: \mathcal{I} \longrightarrow \mathcal{U}$ as follows:

\begin{equation}\label{eq:operator_learning}
    \mathcal{G}(I)(\textbf{x},t) = u(\textbf{x}, t), 
\end{equation}
where $I \in \mathcal{I}, \textbf{x} \in D,$ and $t \in [0, t_{final}]$. So, each initial condition $I \in \mathcal{I}$ is mapped into its corresponding solution $u \in \mathcal{U}$. The goal is to approximate the operator $\mathcal{G}$ using a neural network.

The first SciML operator learning method, called DeepONet, was proposed by Lu et al. \cite{lu2021learning}. The main components of a DeepONet are two neural networks: the branch and the trunk. Each network can be a fully-connected neural network, convolutional, or any other architecture. Usually, the branch inputs are functions, and the trunk inputs are coordinates. DeepONets learn projections from the functions to a vector space, so they can map input functions to output functions at specific points.

Another operator learning approach is the Fourier neural operator (FNO) \cite{li2020fourier, kovachki2021neural}. FNOs, similarly to DeepONets, learn mappings between function spaces using projections. Specifically, FNOs utilize the Fourier transform. They are effective and easy to implement, gaining traction in the SciML community. FNOs are accurate, especially for smooth and periodic problems \cite{lu2022comprehensive}. We note that while the Fourier kernel is continuous, in practice, it is necessary to use discrete versions for operator learning. Consequently, FNOs can be computationally costly when working with high-dimensional problems requiring many Fourier modes.  

\subsubsection{Transformers and attention} 

First presented by Vaswani et al. \cite{vaswani2017attention}, transformers have been widely used in the ML community. Transformers introduce a new type of mechanism called the \textit{scaled dot-product attention}. The attention module attempts to gather context from the given input. It does so by operating on a discrete embedding of the data composed of discrete tokens. 

The original architecture was proposed for natural language processing purposes, where one encodes sentences using their enumerated locations in the vocabulary. Since then, their usage has been extended to many other domains, and they outperform many different deep learning architectures in a wide variety of tasks. These domains include time series analysis \cite{wen2022transformers} and computer vision \cite{han2022survey}. For example, Vision Transformers (ViT) \cite{dosovitskiy2020image} split images into small patches, tokenize them, and apply the attention mechanism. In addition, they are computationally lighter than other alternatives and can be easily parallelized.

Transformers are also becoming increasingly popular in the SciML community. Transformers have been used for operator learning in many different ways \cite{transformerPDE1, transformerPDE2, transformerPDE3, hao2023gnot}. These methods show much promise by using the attention mechanism to find connections between points in the physical domain to function values. Some methods emphasize the attention mechanism itself \cite{transformer_cao_choose, transformer_cao2} and adapt it to PDE-related problems. Others utilize existing transformer models to solve PDE problems more easily \cite{kumar2023crunchgpt}. In this work, we employ elements from the original Transformer architecture as part of the proposed neural network architecture.

\subsubsection{Diffusion models}
\label{supp:DDPM_background}

A diffusion model is a generative deep learning model that uses a Markov chain to produce samples that match a given dataset \cite{sohl2015deep}. These models aim to learn the underlying distribution of a given dataset. After learning this distribution, they are used to generate new samples of similar properties to those found in the training set. 

In \cite{ho2020denoising}, Ho et al. introduced a new type of diffusion model called denoising diffusion probabilistic models (DDPM). It consists of a forward diffusion process and an inverse one. In the forward case, Gaussian noise is incrementally added to the original sample for a given number of iterations. For a sufficiently large number of iterations, the noise completely destroys the original signal. Then, in the reverse diffusion process, the goal is to reconstruct the original signal by performing iterative denoising steps using a neural network. Diffusion models have been used for SciML purposes, especially for generative artificial intelligence purposes \cite{wang2023generative,shu2023physics,liu2023generative}. While we are not using their generative capabilities in this work, we briefly explain their standard training procedure.

We present a mathematical formulation mostly based on the works of Ho et al. \cite{ho2020denoising} and Nichol et al. \cite{nichol2021improved}. Given a data distribution $x_0 \sim q(x_0)$, we define a forward noising process $q$ which produces steps $x_1, \ldots, x_T$ by adding Gaussian noise at time $t$ with variance $\beta_t \in (0, 1)$ as follows:

\begin{equation}\label{markov_diffusion}
    q(x_1, \ldots, x_T | x_0) := \prod_{t=1}^{T} q(x_t | x_{t-1}) ,
\end{equation}

\begin{equation}\label{markov_denoising}
    q(x_t | x_{t-1}) := \mathcal{N}(x_t ; \sqrt{1 - \beta_t} x_{t-1}, \beta_t \textbf{I}) .
\end{equation}

Given a sufficiently large $T$ and a well-behaved schedule $\beta_t$, the latent $x_T$ is nearly an isotropic Gaussian distribution. From \eqref{markov_denoising}, we see that $x_t$ is drawn from a conditional Gaussian distribution with mean $\mu_t = \sqrt{1 - \beta_t} x_{t-1}$ and variance $\sigma^2_t = \beta_t$. In practice, this is done by randomly sampling a noise level parameter $\varepsilon \sim \mathcal{N}(\textbf{0}, \textbf{I})$, and setting:

\begin{equation}\label{diffusion_step}
    x_t = \sqrt{1 - \beta_t} x_{t-1} + \sqrt{\beta_t} \varepsilon .   
\end{equation}

Thus, if we know the exact reverse distribution $q(x_{t-1} | x_t)$, we can sample $x_T \sim \mathcal{N} (0, \textbf{I})$ and run the process in reverse to get a sample from $q(x_0)$. However, since $q(x_{t-1} | x_t)$ depends on the entire data distribution, we approximate it
using a neural network with hyperparameters $\theta$ as follows:

\begin{equation}\label{markov_denoising_nn}
    p_{\theta}(x_{t-1} | x_t) := \mathcal{N}(x_{t-1} ; \mu_{\theta}(x_t, t), \Sigma_{\theta}(x_t, t)).
\end{equation}

The neural network needs to learn the mean and variance to complete the backward diffusion process. Importantly, using the formulation in \eqref{diffusion_step}, in each step, it is sufficient to know $\beta_t, x_t$, and $\varepsilon$ to approximate $x_{t-1}$. Then, the network is used autoregressively to reconstruct $x_0$. Assuming we know the schedules $\{ \beta_t \}_{t=1}^T$, we can view the neural network as the following mapping:

\begin{equation}\label{diffusion_mapping}
    (x_t, \varepsilon) \longrightarrow x_{t-1} .
\end{equation}

In each step, the neural network performs a denoising operation, mapping $x_t$ to a slightly less noisy signal $x_{t-1}$. Including the noise level parameter $\varepsilon$ is essential for the denoising operation. During training, various noise levels are sampled, and knowing the specific noise level that distinguishes between consecutive states $x_t$ and $x_{t-1}$, is crucial for effective denoising. Without this explicit knowledge of the noise level, the denoising process would become significantly more complicated, and the network may not converge. This means we have a conditional denoising operation, conditioned on $\varepsilon$ (or equivalently on the timestep with $\beta_t$).

\subsection{DiTTO: Diffusion-inspired temporal transformer operator}

As described in \nameref{sec:results}, we propose using temporal conditioning, instead of conditioning on noise as in \Cref{supp:DDPM_background}. We use the initial condition of a differential equation, and infer the entire temporal process. It operates on various initial conditions, hence it is performing operator learning according to \Cref{supp:operator_learning}. In addition, after training, the inference is possible on any real temporal value. Hence, the inference is continuous in time. As shown in the results, not only interpolation in time is possible, but also extrapolation.

To train the DiTTO network we gather data of multiple time-dependent procedures, varying in the initial condition. Then, we use the network architecture in \nameref{sec:results}, and select a training strategy (for example, temporal-bundling), to fit the data. We elaborate on these steps in the following subsections.

\subsubsection{Training dataset}
\label{subsubsec:general_dataset}

To train a neural network using the formulation presented in \nameref{sec:results}, we require a large set of initial conditions (inputs) and corresponding solutions (outputs). Let $\{ I^m(\textbf{x}) \}_{m=1}^M$ be a set of initial conditions with corresponding analytic solutions $\{ u^m(\textbf{x},t) \}_{m=1}^M$, where $M$ is the desired number of training samples. Each sample consists of an initial condition and a PDE solution at the relevant timesteps. In practice, $\{ u^m(\textbf{x},t) \}_{m=1}^M$ are numerical approximations of the analytic solutions and not analytic solutions which are often unavailable. Furthermore, the solutions are discretized in space using a grid that partitions the domain $D$. We emphasize that for all $m=1, \ldots, M$ and $t=0,\ldots, T$, $u^m(\textbf{x},t)$ is a matrix, and its dimensions depend on the spatial discretization parameters, i.e., the number of nodes along each axis.

We denote the forward process corresponding to the $m$-th initial condition and solution by $\{ x_{n}^m \}_{n=0}^T$, where $x_{n}^m := u^m(\textbf{x},t_n)$. We define the following datasets:

\begin{equation}\label{dataset}
   \begin{aligned}
    \textbf{X} = \{ ( x_{0}^m, t_n ) | \; n=1, \ldots, T, \;\;\; m=1, \ldots, M\} \\
    \textbf{Y} = \{ x_{n}^m | \; n=1, \ldots, T, \;\;\; m=1, \ldots, M\} 
    \end{aligned}.
\end{equation} 

So, each solution of the PDE is transformed into $T$ pairs of samples that correspond to the mapping described in \Cref{eq:ditto_operaor_mapping}.

\subsubsection{The DiTTO neural network architecture and parameters}

We use the architecture described in \nameref{sec:results}. We mention that for the temporal input $t$, we use an embedding mechanism based on the original Transformer positional encoding \cite{vaswani2017attention}:

\begin{equation}\label{eq:temporal_embedding}
\begin{aligned}
        PE_{(pos, 2i)} = \sin{\left(\frac{pos}{10000^{\sfrac{2i}{d_{emb}}}}\right)}, \\ 
        PE_{(pos, 2i + 1)} = \cos{\left(\frac{pos}{10000^{\sfrac{2i}{d_{emb}}}}\right)}, \\ 
\end{aligned}
\end{equation}
where $d_{emb}$ is the desired embedding dimension. Each scalar $t$ is mapped into a vector of size $d_{emb}$. 

We now describe the loss function used as a target for training the network. Let $\mathcal{O}_\theta$ be the neural network described in \nameref{sec:results} with hyperparameters $\theta$. The goal of $\mathcal{O}_\theta$ is to learn the mapping described in \eqref{eq:ditto_operaor_mapping}, using the dataset \eqref{dataset}. We split this dataset into training, validation, and testing sets. We split them in a way that makes sure that no initial conditions from the validation and testing sets appear in the training set.

Diffusion models are often trained with a probabilistic loss function. However, since we learn a PDE operator, other loss functions commonly used for SciML applications are more fitting. Consequently, we train the network with a mean relative $\textit{L}^2$ loss:

\begin{align}\label{loss_function}
    loss :=\frac{1}{M T} 
    \sum_{m=1}^{M}\sum_{n=1}^T
    \frac{||\mathcal{O}_\theta(x_{0}^m, t_n) - x^m_n||_2}{\varepsilon+||x^m_n||_2},
\end{align}

where $\varepsilon$ is a small number used to prevent a zero denominator and stabilize the loss. The inputs and outputs of the model are $d$-dimensional, so they are converted into a one-dimensional array by column stacking (flattening) inside the loss function when needed. We describe the loss for the entire dataset for simplicity, but in practice, we divide it into batches.

Iterating over the entire dataset \eqref{dataset} can be time-consuming. For $M_{train}$ initial conditions in the training set, we have $M_{train} \cdot T$ samples. So, the number of training steps scales linearly with $T$. This means the number of training samples is very large for fine temporal discretizations. 

A similar problem occurs in generative diffusion models. The original DDPM \cite{ho2020denoising} requires hundreds of forward passes to produce good results. Later works suggested ways to improve the performance aspect of DDPMs. For example, Song et al. \cite{song2020denoising} suggest using non-Markovian processes to improve the computational cost. Nichol et al. \cite{nichol2021improved} present a way to significantly reduce the number of necessary steps by sub-sampling the original diffusion process. Both methods focus primarily on the inference speed. However, in the case of DiTTO, inference is immediate. In \nameref{sec:results}, we explained that we do not view $x_0, x_1, \ldots, x_T$ as an iterative process. Instead, we treat each sample individually, significantly increasing the inference speed compared to generative models such as DDPM. Hence, we focus on speeding up the training process. 

\subsubsection{DiTTO-s}
\label{subsubsec:ditto-s}

We propose DiTTO-s, a faster variant of DiTTO that relies on a sub-sampling mechanism similar to \cite{nichol2021improved}. Instead of iterating over the entire process, we iterate over a random subsequence. Recall that for the $m$-th initial condition in the training set, the full process is $\{ x_n^{m} \}_{n=1}^{T}$. Instead, we take a set of random subsequences $S_m \subset \{0, 1, \ldots, T \}$, such that $\sum_{m=1}^M |S_m| = \alpha M T$ for some $\alpha < 1$. For example, choosing $\alpha = 0.05$ means we only use $5 \%$ of the given samples in each epoch. The new DiTTO-s loss is given by: 

\begin{align}\label{loss_function_s}
    loss_{\alpha} := \frac{1}{\alpha M T} 
    \sum_{m=1}^{M}\sum_{n \in S_m}
    \frac{||\mathcal{O}_\theta(x_{0}^m, t_n) - x^m_n||_2}{\varepsilon+||x^m_n||_2},
\end{align}

After each epoch, we randomly sample $S_m$ again using a uniform distribution. That way, given a sufficiently large number of epochs, we expect to cover a significant portion of samples in the dataset.

\subsubsection{DiTTO-point}
\label{subsubsec:ditto-point}

The architecture shown in \Cref{fig:ditto_architecture} can be used for problems in different dimensions. One way to accomplish that is to use $d$-dimensional convolutions in all the convolutional layers, where $d \in \{1,2,3\}$ is the physical dimension of the problem. However, using high-dimensional convolutions requires a large amount of memory. Furthermore, many physical problems are not defined on structured grids. Thus, in order to use a neural operator framework, it is often necessary to project their solutions onto regular grids \cite{lu2022comprehensive}, which requires a change to the geometry of the problem.  

To address these issues, we propose DiTTO-point, another variant of DiTTO that solves high-dimensional problems using exclusively 1-D convolutions. With DiTTO-point we treat the domain as a set of points in space instead of a structured $d$-dimensional grid. A domain with $N$ points is defined as a $N \times d$ matrix, where each row represents a spatial coordinate. Similarly, we define the solution on this domain using a vector of size $N$, corresponding to the values of the solution at each point of the domain. Using this formulation, regardless of the original dimensionality of the problem, the input to DiTTO-point would always be of size $N \times d$. This enables the use of 1-D convolutions on this data, where $d$ is the number of input channels.

However, directly using the architecture shown in \Cref{fig:ditto_architecture} with 1-D convolution does not work, without any modifications, on high-dimensional problems. Importantly, switching from a structured grid to a set of coordinates results in the loss of spatial information. This is especially true when the order in which the coordinates appear is not necessarily related to their physical distance. To solve this issue, we use another layer of positional encoding (see \Cref{eq:temporal_embedding}) based on the spatial coordinates of the grid. We apply this layer to the beginning of the overall architecture before we concatenate its output with the relevant initial condition in the latent space. This enables DiTTO-point to keep high-dimensional spatial information while using a one-dimensional architecture.

\subsubsection{DiTTO-gate}
\label{subsubsec:ditto-gate}

We propose another variant of the architecture shown in \Cref{fig:ditto_architecture}, called DiTTO-gate. In DiTTO-gate we modify the behavior of the skip connections of the U-Net decoder. An analysis of diffusion models shows that the U-Net skip connections introduce high-frequency features into the decoder \cite{si2023freeu}. So, for scenarios with fine details and sharp features (such as Navier-Stokes with high Reynolds numbers), we put extra emphasis on the skip connections. So, we introduce a gate component, which operates directly on the skip connections. This component is composed of a standard convolutional block, as described in \nameref{sec:results}. So, in DiTTO-gate, we add such a component to each level of the decoder and use it on its incoming skip connection. 

\FloatBarrier
\section{Hypersonic flow data generation and details}\label{app:hypersonics}

We first generate a training dataset by solving the 2D Euler equations on a fluid domain around a double-cone geometry. The governing equations read

\begin{equation}
\frac{\partial \mathbf{U}}{\partial t}+\frac{\partial \mathbf{F} }{\partial x}+\frac{\partial \mathbf{G} }{\partial y}=\mathbf{0},
\label{eq:vec_gov}
\end{equation}
where the vector of conservative variables, x-direction and y-direction flux vectors, are described as

\begin{equation}
\mathbf{U}=\begin{pmatrix}\rho\\ \rho u \\ \rho v\\ \rho E
\end{pmatrix}, \quad \mathbf{F}=\begin{pmatrix}\rho u\\ \rho u^2+p \\\rho uv\\ u(\rho E+p)
\end{pmatrix},\quad \mathbf{G}=\begin{pmatrix}\rho v\\ \rho vu\\\rho v^2+p \\ u(\rho E+p)
\end{pmatrix}.
\label{eq:vec}
\end{equation}
In \Cref{eq:vec_gov}, $t$ is time, $x$, $y$ denote the spatial coordinates, $u$ and $v$ indicate x-direction and y-direction velocities, $\rho$ is density, and $p$ represents the pressure. The total energy in \Cref{eq:vec} is illustrated as 

\begin{equation}
   \rho E = \frac{p}{\gamma-1} +\frac{1}{2}\rho (u^2+v^2),
    \label{viscous}
\end{equation}
where $\gamma$ is the ratio of specific heats that is assumed constant with a value of $1.4$.

\begin{figure}[h]
    \centering
    \includegraphics[scale=0.4]{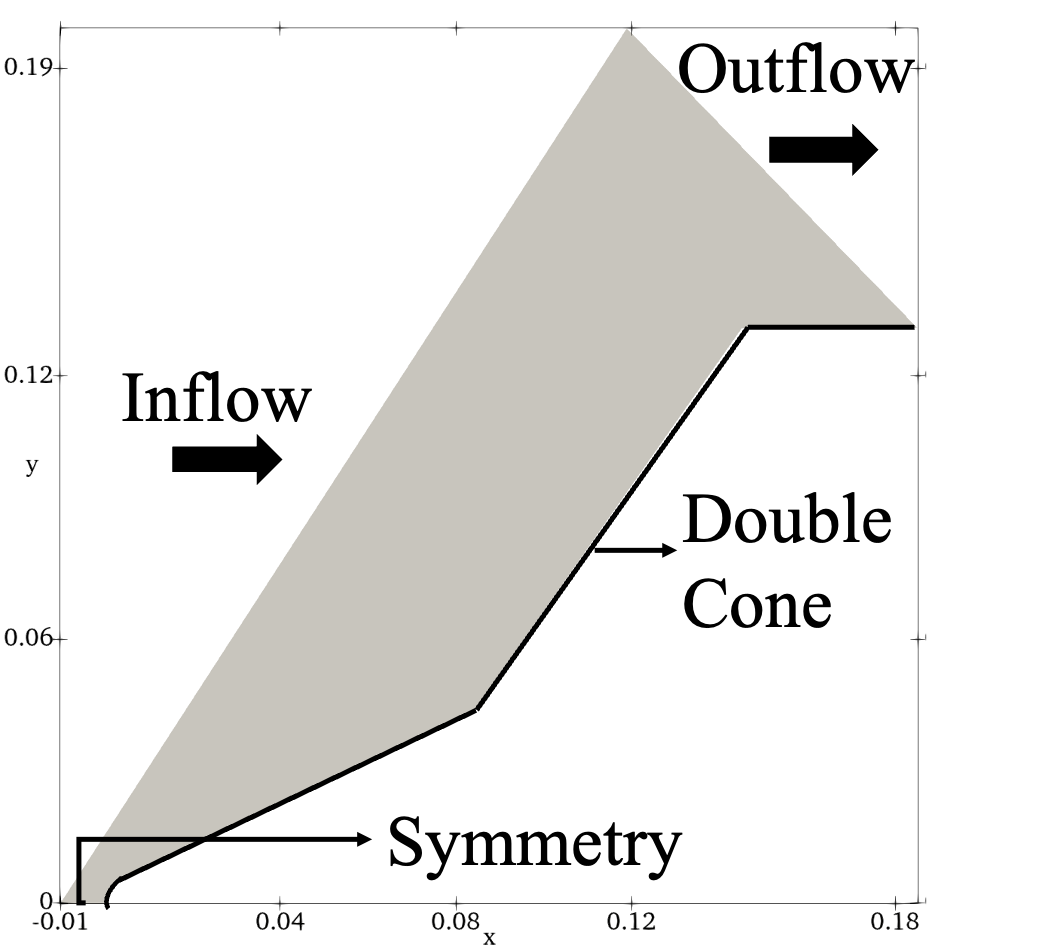}
    \caption{The 2D domain of the double cone problem. The gray area is the domain and the highlighted lines show the surface of the double cone which is considered as a slip wall boundary condition. The small line at the bottom is defined as a symmetry condition and the Inflow and outflow boundaries are also shown.}
    \label{fig:double_cone}
\end{figure}

We solve the system of equations using Trixi.jl numerical framework \cite{schlottkelakemper2020trixi}, which employs entropy stable discontinuous Galerkin spectral element (ES-DGSEM) approach \cite{ranocha2022adaptive,schlottkelakemper2021purely, PEYVAN2023112310} to solve hyperbolic and parabolic systems of equations. The ES-DGSEM features high accuracy and stability and also employs adaptive mesh refinement to adapt the mesh resolution automatically to the high gradient regions of the flow field. The 2D physical domain is shown in \Cref{fig:double_cone}, where the boundary conditions are specified. The domain is initialized using constant uniform values for primitive variables as $u=M$, $v=0$, $p=1.0$, and $\rho=1.4$. We initialize the x-direction velocity with the value of the free-stream Mach number denoted by $M$. Each simulation is performed for a time span of $t\in [0,0.04]$, and the solution values are stored at $201$ snapshots corresponding to equidistance instances of time.

\begin{table}[!htb]
\centering 
\begin{tabular}{|l|cccccc|}
\hline
 \backslashbox{Model}{Noise level} & $0 \%$ & $10 \%$ & $20 \%$ & $30 \%$ & $50 \%$ & $100 \%$ \\
\hline
\hline
DiTTO & 0.0303 & 0.0360 & 0.0464 & 0.0591 & 0.0921 & 0.2030 \\
\hline
DiTTO-point & 0.0152 & 0.0155 & 0.0167 & 0.0188 & 0.0253 & 0.0540 \\
\hline
FNO & 0.0241 & 0.0795 & 0.1519 & 0.2229 & 0.3544 & 0.6080 \\
\hline
\end{tabular}
\caption{Relative $\textit{L}^2$ test set errors for the hypersonic flow case \nameref{subsubsec:hypersonics} for different noise levels.}
\label{tab:noise}
\end{table}

\FloatBarrier
\section{Additional results}\label{app:additional_results}

\begin{table}[!htb]
\centering 
\begin{tabular}{|ll|ccccc|}
\hline 
 Scenario & Model & $N_t^{test} = 10$ & $N_t^{test} = 20$ & $N_t^{test} = 50$ & $N_t^{test} = 100$ & $N_t^{test} = 200$ \\
\hline\hline
\multirow{4}{9em}{1D Burgers' \\ $t_{final}=1,  \nu = 0.01$} & DiTTO & 0.0060 & 0.0059 & 0.0058 & 0.0065 & 0.0070 \\
& DiTTO-s & \textbf{0.0057} & \textbf{0.0056} & \textbf{0.0055} & \textbf{0.0061} & \textbf{0.0066} \\
& FNO & 0.1672 & 0.0217 & 0.0059 & 0.0100 & 0.0124 \\
& U-Net & 0.4096 & 0.1668 & 0.0094 & 0.1017 & 0.1782 \\
\hline
\multirow{4}{9em}{1D Burgers' \\ $t_{final}=1,  \nu = 0.001$} & DiTTO   & 0.0164 & 0.0157 & 0.0154 & 0.0153 & 0.0152 \\
& DiTTO-s & \textbf{0.0124} & \textbf{0.0119} & \textbf{0.0116} & \textbf{0.0116} &\textbf{0.0116} \\
& FNO     & 0.2899 & 0.0461 & 0.0222 & 0.0265 & 0.0299 \\
& U-Net    & 0.4789 & 0.2807 & 0.0298 & 0.2042 & 0.3188 \\
\hline
\multirow{4}{9em}{1D Burgers' \\ $t_{final}=2,  \nu = 0.001$} & DiTTO & 0.0208 & 0.0199 & 0.0195 & 0.0195 & 0.0196 \\
& DiTTO-s & \textbf{0.0172} & \textbf{0.0164} & \textbf{0.0160} & \textbf{0.0160} & \textbf{0.0161} \\
& FNO & 0.3501 & 0.0582 & 0.0248 & 0.0315 & 0.0367 \\
& U-Net & 0.6655 & 0.3795 & 0.0382 & 0.2426 & 0.3167 \\
\hline
\multirow{4}{9em}{2D Navier Stokes \\ $Re \approx 20$} 
& DiTTO               &    \textbf{0.2948} &             0.1534 &             0.0711 &              0.0465 &              0.0381 \\
& DiTTO-point         &             0.2982 &             0.1540 &             0.0668 &     \textbf{0.0401} &     \textbf{0.0291} \\
& DiTTO-s             &             0.2973 &             0.1553 &             0.0715 &              0.0475 &              0.0392 \\
& DiTTO-point-s       &             0.3062 &             0.1589 &             0.0694 &              0.0418 &              0.0307 \\
& DiTTO-point-s-gate  &             0.2964 &             0.1551 &             0.0707 &              0.0453 &              0.0359 \\
& FNO                 &             0.3230 &    \textbf{0.1411} &             0.0523 &              0.0690 &              0.0848 \\
& UNet                &             0.6467 &             0.8332 &    \textbf{0.0489} &              0.6361 &              0.6154 \\
\hline
\multirow{4}{9em}{2D Navier Stokes \\ $Re \approx 2,000,$ \\ $\#samples=1,000$} 
& DiTTO               &             0.2491 &             0.1987 &             0.1745 &              0.1685 &              0.1660 \\
& DiTTO-point         &             0.2344 &             0.1813 &             0.1550 &              0.1485 &              0.1458 \\
& DiTTO-s             &             0.2414 &             0.1910 &             0.1669 &              0.1610 &              0.1586 \\
& DiTTO-point-s       &             0.2305 &             0.1764 &             0.1505 &              0.1443 &              0.1418 \\
& DiTTO-point-s-gate  &    \textbf{0.2187} &    \textbf{0.1646} &    \textbf{0.1381} &     \textbf{0.1317} &     \textbf{0.1291} \\
& FNO                 &             0.3560 &             0.2040 &             0.1617 &              0.1651 &              0.1700 \\
& UNet                &             0.7033 &             0.6723 &             0.1955 &              0.6331 &              0.7385 \\
\bottomrule
\end{tabular}
\caption{A full error analysis of all cases comparing the various DiTTO methods with FNO and U-Net across different temporal resolutions.}
\label{tab:app_full_comparison}
\end{table}

\subsection{One-dimensional Burgers' Equation}
\label{app:burgers}

The one-dimensional time-dependent Burgers' equation for a viscous fluid is given by:

\begin{equation}\label{burgers_equation}
\begin{cases}
    \partial_t u(x,t) + \partial_x (u^2(x,t) / 2) = \nu \partial_{xx} u(x,t), & x \in (0,1), t \in (0, t_{final}] \\
    u(x,0) = u_0, & x \in (0,1) \\
\end{cases},
\end{equation}
where $\nu \in \mathbb{R}^+$ is the viscosity coefficient and is subject to periodic boundary conditions. The initial condition $u_0(x)$ is sampled from a Gaussian random field according to the following distribution: $\mathcal{N}(0,625 (-\Delta + 25 I)^{-2})$, where $\mathcal{N}$ is the normal distribution, and $\Delta, I$ are the Laplacian and identity operators, respectively. 

We use the publicly available MATLAB \cite{MATLAB} solver given in \cite{li2020fourier} to create three separate datasets with different parameters. The first dataset is created with $\nu = 0.01, t_{final}=1,$ and $N_x = 128$. This is a relatively simple scenario since these parameters produce smooth solutions without shocks. In the second dataset, we decrease the viscosity coefficient to $\nu = 0.001$, which increases the shock-like behavior of the Burgers' equation. For this reason, we also increase the spatial discretization to $N_x = 256$. Finally, the third dataset is the same as the second one, except we increase the final simulation time to $t_{final} = 2$, which causes the shocks to be more pronounced.

The results for the three scenarios are shown in \Cref{tab:full_comparison}. DiTTO and DiTTO-s achieve the lowest errors for the three datasets. When $N_t^{test} = N_t^{train}=50$, all methods have similar errors, with DiTTO and DiTTO-s having a slight advantage. However, when $N_t^{test} \neq N_t^{train}$, DiTTO and DiTTO-s significantly outperform the FNO and the U-Net. Moreover, we observe that the DiTTO and DiTTO-s results do not depend on the temporal discretization, as the errors stay roughly the same for all values of $N_t^{test}$. We also note that DiTTO-s has a slightly lower error than the full DiTTO. This demonstrates that the sub-sampling mechanism does not only require fewer training steps but also improves the model. The reason for this is that the sub-sampling mechanism acts as a form of regularization that helps decrease the error.

\begin{figure}
  \centering
    \begin{subfigure}{16cm} 
     \includegraphics[scale=0.33]{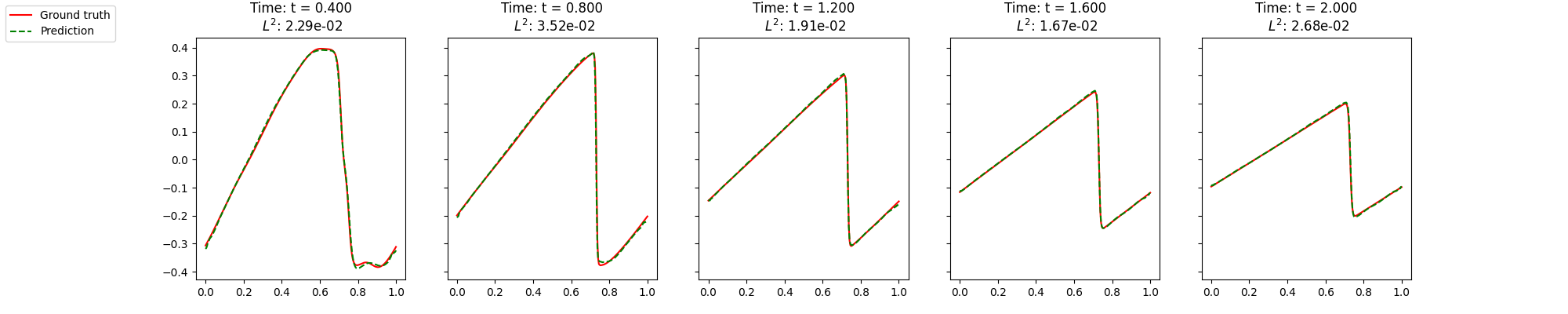}
     \caption{DiTTO-s}
     \label{fig:burgers_ditto_s}
    \end{subfigure}
    \begin{subfigure}{16cm} 
     \includegraphics[scale=0.33]{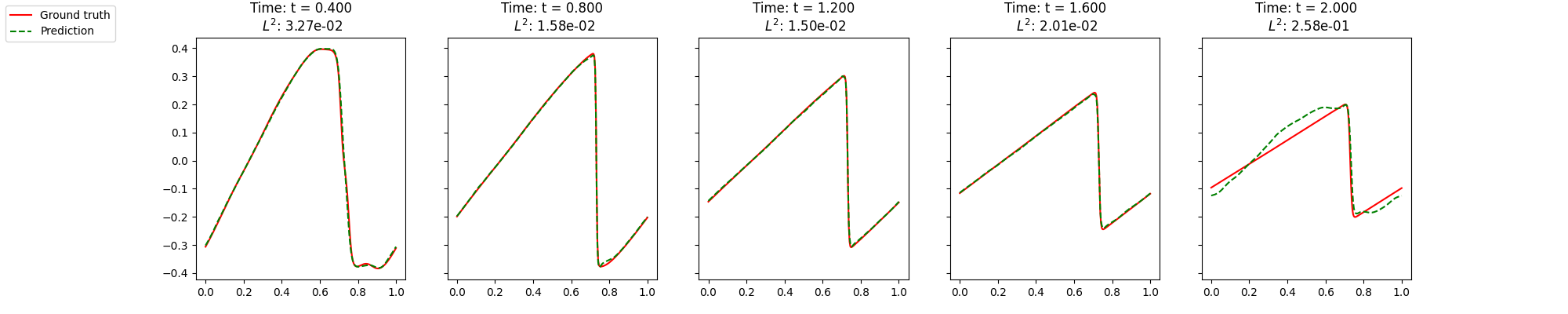}
     \caption{FNO}
     \label{fig:burgers_fno}
    \end{subfigure}
    
    \begin{subfigure}{16cm} 
     \includegraphics[scale=0.33]{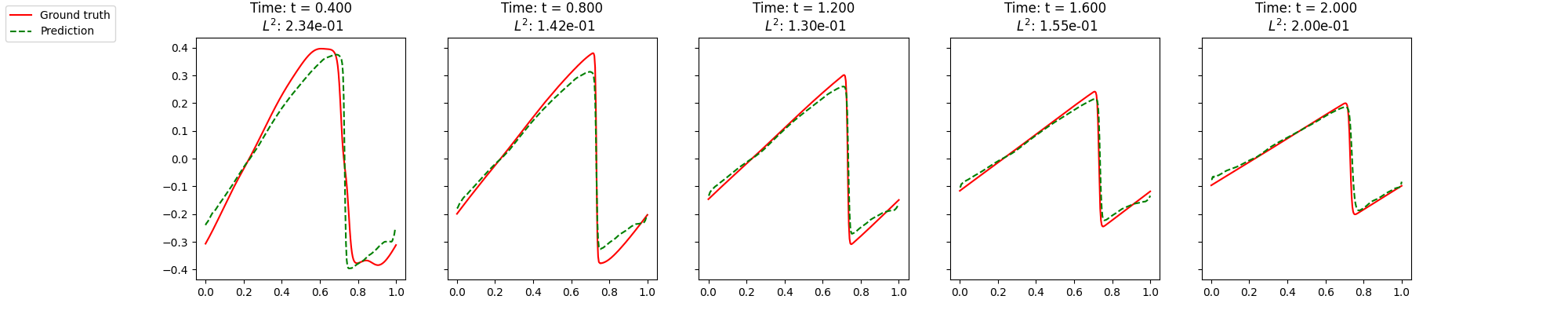}
     \caption{U-Net}
     \label{fig:burgers_U-Net}
    \end{subfigure}
\caption{Results for the Burgers' problem described in \Cref{app:burgers} with $\nu = 0.001, t_{final} = 2, N_x = 256, N_t^{train}=50,$ and $N_t^{test}=200$, for a random initial condition from the test set. In \Cref{fig:burgers_ditto_s,fig:burgers_fno,fig:burgers_U-Net}, we see a comparison between the models at different times. We plot the predictions of the models alongside the reference solution (ground truth).}
\label{fig:burgers}
\end{figure}

\subsection{Two-dimensional Navier-Stokes Equations}
\label{app:ns}

The time-dependent two-dimensional incompressible Navier-Stokes equation for a viscous, incompressible fluid in vorticity form is given by:

\begin{equation}\label{eq:NS_equation}
\begin{cases}
    \partial_t \omega(x,y,t) + u(x,y,t) \cdot \nabla  \omega(x,y,t) = \nu \Delta \omega(x,y,t) + f(x,y), & x,y \in (0,1)^2, t \in (0, t_{final}] \\
    \nabla \cdot u(x,y,t) = 0, & (x,y) \in (0,1)^2, t \in (0, t_{final}] \\
    \omega(x,y,0) = \omega_0, & (x,y) \in (0,1)^2 \\
\end{cases}
\end{equation}
where $\omega$ is the vorticity, $u$ is the velocity field,  $\nu$ is the viscosity, and $\Delta$ is the two-dimensional Laplacian operator. We consider periodic boundary conditions. The source term $f$ is given by $f(x,y) = 0.1 (sin(2\pi (x + y)) + cos(2\pi(x + y)))$, and the initial condition $\omega_0(x)$ is sampled from a Gaussian random field according to the distribution $\mathcal{N}(0, 7^{3/2} (-\Delta + 49 I)^{-5/2})$. 

We use the publicly available Python solver given in \cite{li2020fourier} to generate two datasets with a spatial resolution of $N_x = N_y = 64$. The first dataset is created with $\nu = {10}^{-3}$ and $t_{final} = 50$, resulting in a Reynolds number $Re \approx 20$. For the second dataset we use $\nu = {10}^{-5}$ and $t_{final} = 20$, resulting in a Reynolds number $Re \approx 2,000$. 

The error comparison for the two datasets is shown in \Cref{tab:full_comparison}. For $Re \approx 20$, both DiTTO and DiTTO-s outperform the other models across all temporal discretizations while keeping similar error values. For $Re \approx 2,000$, it is clear that increasing the Reynolds number also increases the difficulty of the problem, as evidenced by the noticeably higher errors for all models. We also see that the FNO errors are closer to DiTTO-s compared to other cases, even having a slight advantage when $N_t^{test} \in \{50, 100\}$. 


\begin{figure}
    \centering
    \begin{subfigure}{18cm}  
     \includegraphics[scale=0.33]{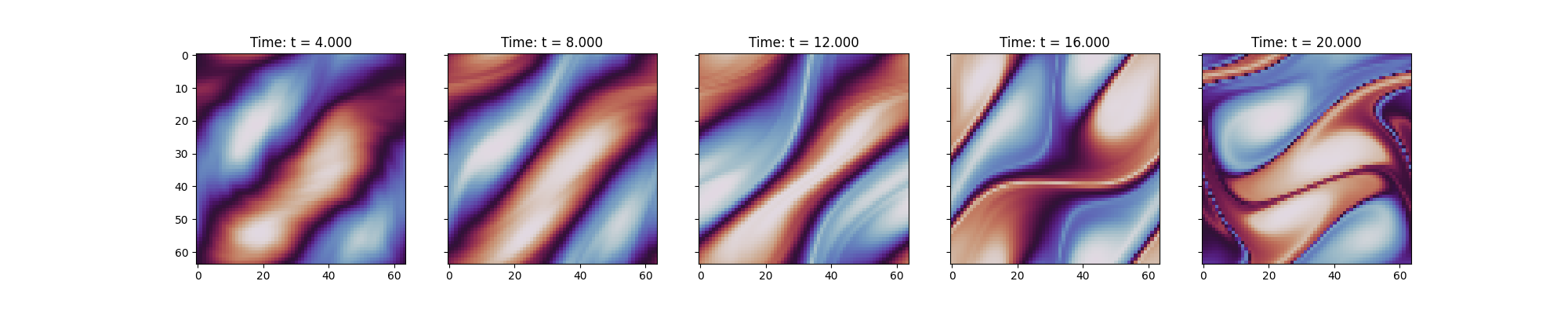}
     \caption{Reference solution}
     \label{fig:app1_nx_re_2000_ref}
    \end{subfigure}
    
    \begin{subfigure}{18cm} 
     \includegraphics[scale=0.33]{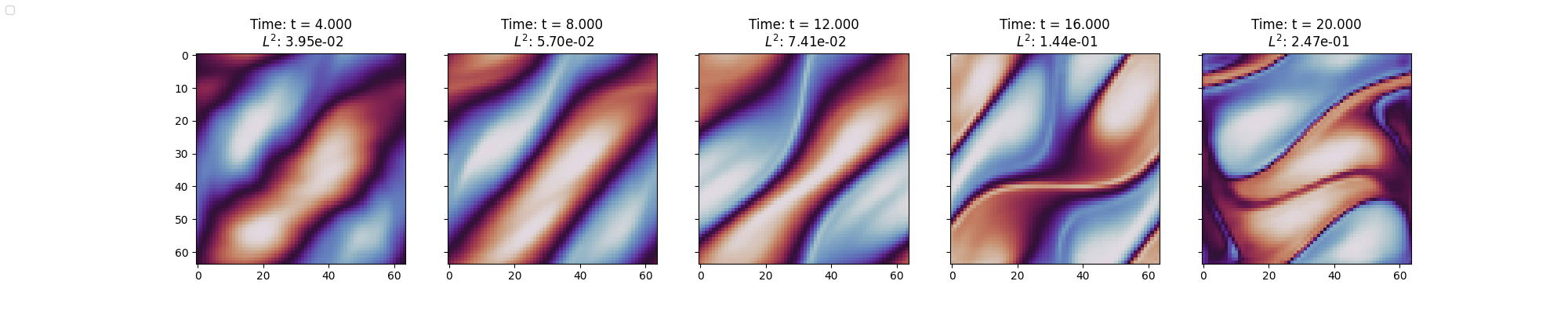}
     \caption{DiTTO}
     \label{fig:app1_nx_re_2000_ditto}
    \end{subfigure}
    
    \begin{subfigure}{18cm} 
     \includegraphics[scale=0.33]{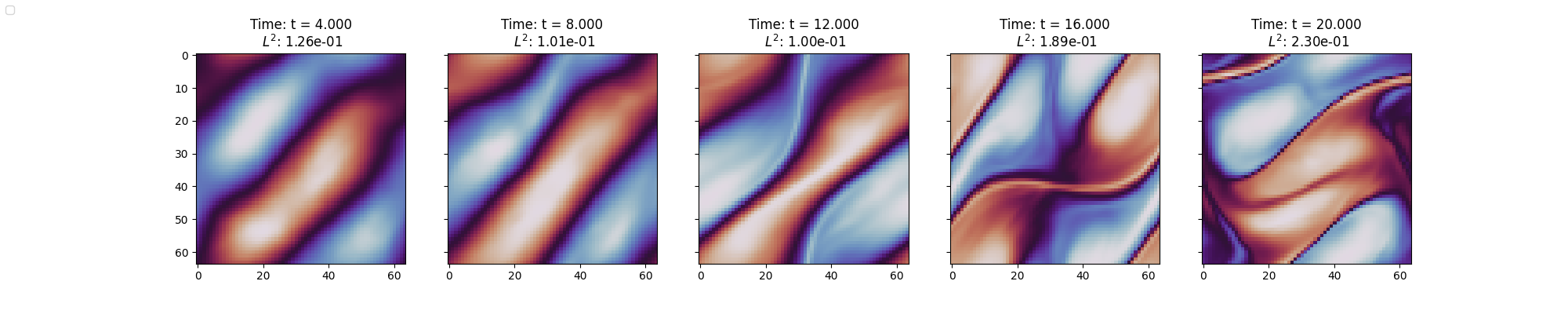}
     \caption{FNO}
     \label{fig:app1_nx_re_2000_fno}
    \end{subfigure}
    
    \begin{subfigure}{18cm} 
     \includegraphics[scale=0.33]{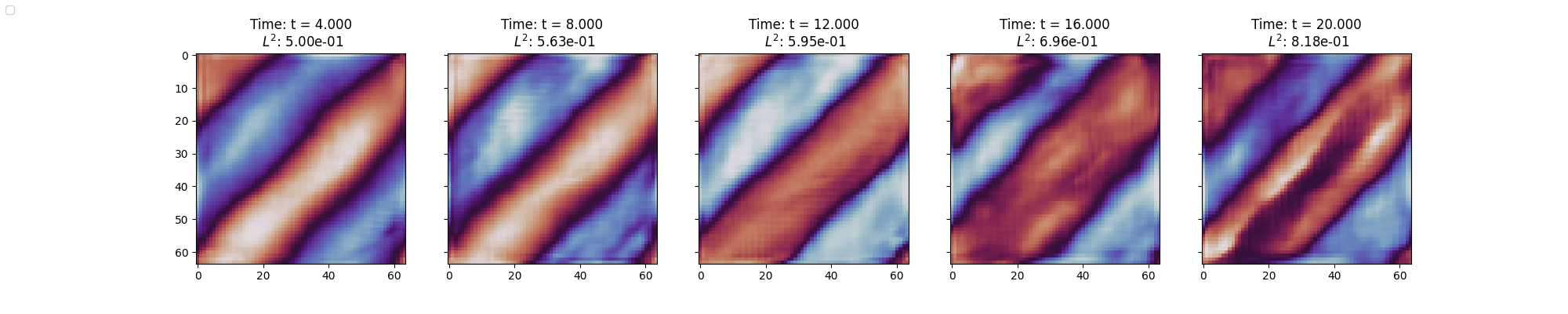}
     \caption{UNet}
     \label{fig:app1_nx_re_2000_unet}
    \end{subfigure}
\caption{Navier-Stokes equations \Cref{eq:NS_equation} with $Re \approx 2,000$.}
\label{fig:app1_ns_re_2000}
\end{figure}

\begin{figure}
    \centering
    \begin{subfigure}{18cm}  
     \includegraphics[scale=0.33]{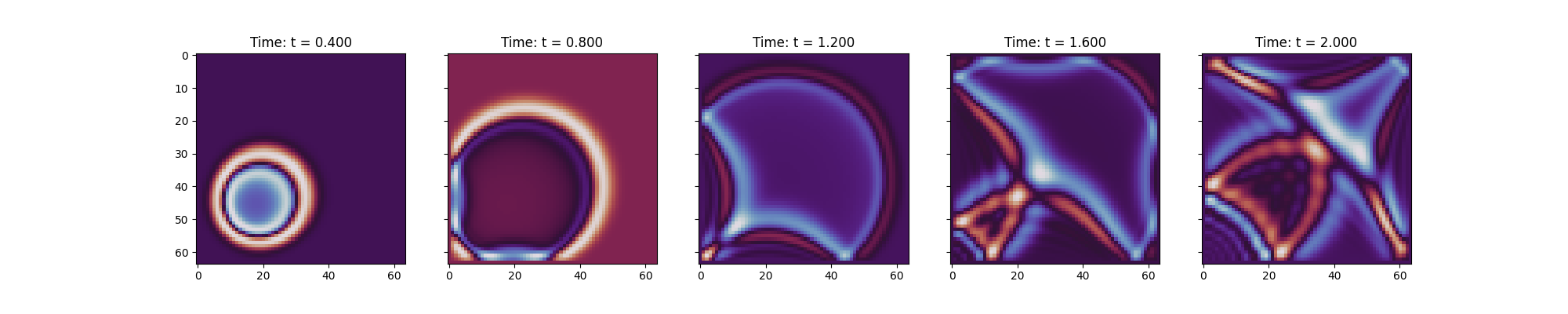}
     \caption{Reference solution}
     \label{fig:app1_wave_eq_2d_ref}
    \end{subfigure}
    
    \begin{subfigure}{18cm} 
     \includegraphics[scale=0.33]{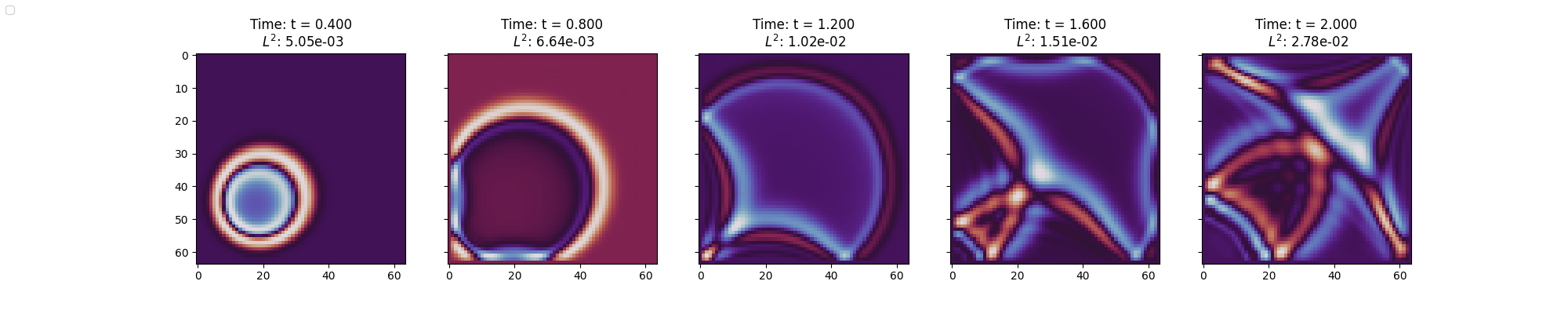}
     \caption{DiTTO}
     \label{fig:app1_wave_eq_2d_ditto}
    \end{subfigure}
    
    \begin{subfigure}{18cm} 
     \includegraphics[scale=0.33]{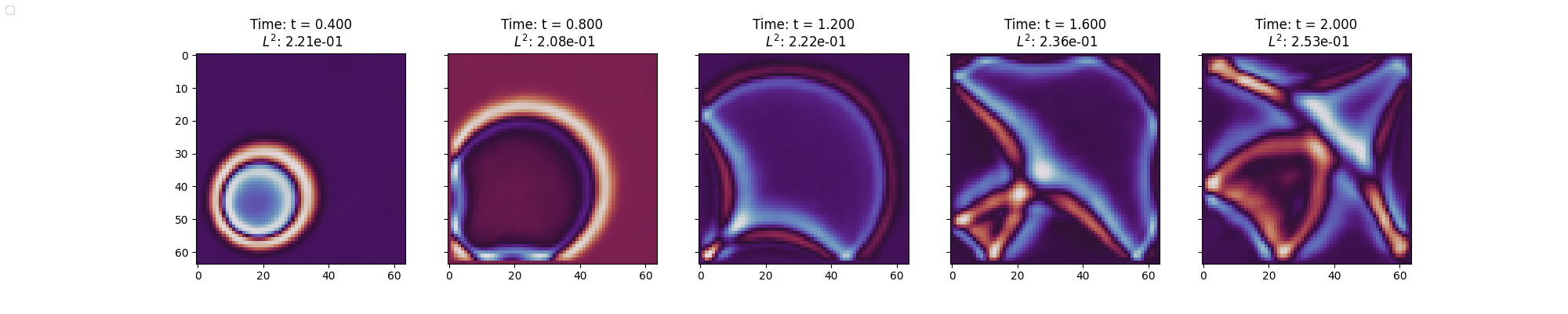}
     \caption{FNO}
     \label{fig:app1_wave_eq_2d_fno}
    \end{subfigure}
    
    \begin{subfigure}{18cm} 
     \includegraphics[scale=0.33]{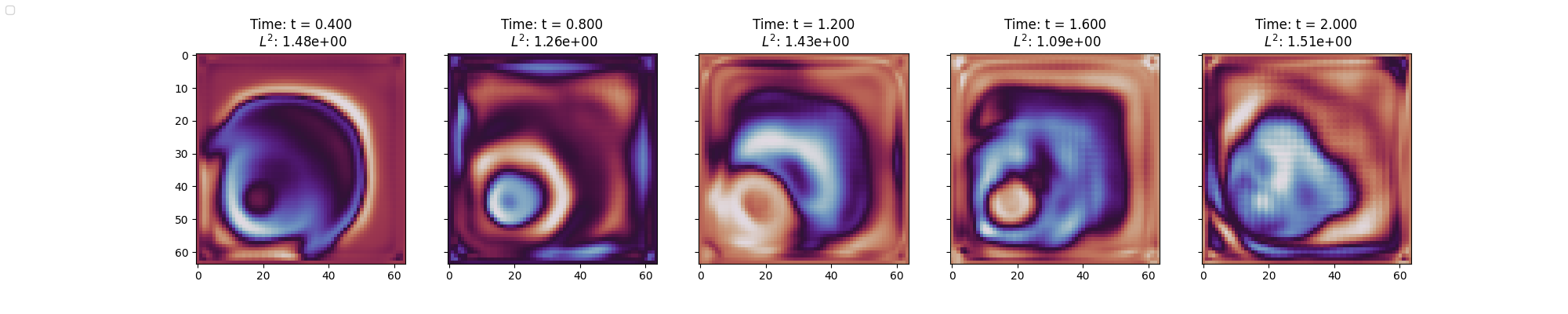}
     \caption{UNet}
     \label{fig:app1_wave_eq_2d_unet}
    \end{subfigure}
\caption{2D Acoustic wave equation \ref{eq:gen_wave_eq}.}
\label{fig:app1_wave_eq_2d}
\end{figure}

\FloatBarrier
\section{Ablation study}\label{app:ablation}
Here we present an ablation study related to the results shown in  \nameref{sec:results}. Due to the computational cost of many experiments, we conducted the ablation study on one fixed scenario. Specifically, we used the Burgers' dataset described in \Cref{app:burgers} with $\nu = 0.01, T=1, N_x=128,$ and $N_t^{train}=50$.

\textbf{Influence of attention} We examined the impact of the attention mechanism. To assess its contribution, we trained two models and compared them. The first model was DiTTO, as described in \nameref{sec:results}. The second model was the same, except we removed all attention layers.  The full results are shown in \Cref{tab:ablation_attention}. It is clear that the attention mechanism roughly halved the error values across all choices of $N_t$.

\begin{table}[!htb]
\centering 
\begin{tabular}{cccccc}
\toprule
 & $N_t^{test} = 10$ & $N_t^{test} = 20$ & $N_t^{test} = 50$ & $N_t^{test} = 100$ & $N_t^{test} = 200$ \\
\midrule
DiTTO & 0.0060 & 0.0059 & 0.0058 & 0.0065 & 0.0070 \\
DiTTO (no attention) & 0.0132 & 0.0128 & 0.0126 & 0.0127 & 0.0128 \\
\bottomrule
\end{tabular}
\caption{Relative $\textit{L}^2$ test set errors with and without the attention mechanism for the Burgers' scenario in \Cref{app:burgers} with $t_{final}=1, \; \nu = 0.01, \; N_x = 128, \: N_t^{train}=50$.}
\label{tab:ablation_attention}
\end{table}

\textbf{Effect of sub-sampling rate} We experiment with different values of $\alpha$ as described in \Cref{subsubsec:ditto-s} and train several DiTTO-s models with corresponding sub-sampling rates. The results for $\alpha \in \{0.05, 0.1, 0.2, 1 \}$ are shown in \Cref{tab:ablation_alpha}. We see that using the full data in each batch (i.e., $\alpha = 1$) does not necessarily produce the best results. Choosing $\alpha = 0.1$ and $\alpha = 0.2$ produced lower errors compared to $\alpha = 1$. However, $\alpha = 0.05$ was too small and consequently increased the error. Hence, choosing $\alpha = 0.1$ both improved the results and effectively reduced the batch size by $90 \%$, and thus reduced the memory footprint of the model.

\begin{table}[!htb]
\centering 
\begin{tabular}{c|cccccc}
\toprule
 $\alpha$ & $N_t^{test} = 10$ & $N_t^{test} = 20$ & $N_t^{test} = 50$ & $N_t^{test} = 100$ & $N_t^{test} = 200$ \\
\midrule
1    & 0.0060 & 0.0059 & 0.0058 & 0.0065 & 0.0070 \\
0.2  & 0.0058 & 0.0056 & 0.0056 & 0.0061 & 0.0066 \\
0.1  & 0.0057 & 0.0056 & 0.0055 & 0.0061 & 0.0066 \\
0.05 & 0.0250 & 0.0247 & 0.0247 & 0.0248 & 0.0248 \\
\bottomrule
\end{tabular}
\caption{Relative $\textit{L}^2$ test set errors using different sub-sampling rates $\alpha$ for the Burgers' scenario in \Cref{burgers_equation} with $t_{final}=1, \; \nu = 0.01, \; N_x = 128, \: N_t^{train}=50$. }
\label{tab:ablation_alpha}
\end{table}

\textbf{FNO grid search} To increase the validity of the comparison with FNOs, we conducted a hyperparameter search over various FNO architectures. We focused on the number of modes and the depth of the network, as defined in \cite{li2020fourier}. We tested for $N_t^{test} >= 50$ to avoid padding issues for higher numbers of Fourier modes. The results are presented in \Cref{tab:ablation_fno}. While wider, deeper FNO models generally obtained better results for $N_t^{test} = N_t^{train}$, we found that 4 layers with 18 Fourier modes gave the best results in the zero-shot super-resolution case. We followed a similar process for FNO-3D and decided to use 4 layers with 12 Fourier modes for the two-dimensional time-dependent problems.

\begin{table}[!htb]
\centering 
\begin{tabular}{cc|ccc}
\toprule
$\#$ Fourier modes & $\#$ Layers & $N_t^{test} = 50$ & $N_t^{test} = 100$ & $N_t^{test} = 200$ \\
\midrule
12 & 4 & 0.0078          & 0.0112 & 0.0131 \\
18 & 4 & 0.0048          &\textbf{ 0.0080 }& \textbf{0.0100} \\
24 & 4 & 0.0046          & 0.0146 & 0.0184 \\
30 & 4 & \textbf{0.0041} & 0.0262 & 0.0312 \\
12 & 8 & 0.0067          & 0.0087 & 0.0104 \\
18 & 8 & 0.0052          & 0.0205 & 0.0274 \\
24 & 8 & 0.0046          & 0.0162 & 0.0217 \\
30 & 8 & \textbf{0.0041} & 0.0250 & 0.0292 \\
\bottomrule
\end{tabular}
\caption{Relative $\textit{L}^2$ test set errors using different FNO hyperparameter choices for the Burgers' scenario in \Cref{app:burgers} with $t_{final}=1, \; \nu = 0.01, \; N_x = 128, \: N_t^{train}=50$. }
\label{tab:ablation_fno}
\end{table}

\end{appendices}

\end{document}